\documentclass[letterpaper, 10 pt, conference]{ieeeconf}  
\IEEEoverridecommandlockouts                             
\usepackage{graphics} 
\usepackage{epsfig} 
\usepackage{times} 
\usepackage{amsmath} 
\usepackage{amssymb}  
\usepackage{xcolor}
\usepackage{bm}
\usepackage{cancel}
\usepackage{color}
\usepackage{setspace}
\usepackage{wrapfig}
\usepackage{tikz}
\usetikzlibrary{shapes,arrows.meta,calc}
\usetikzlibrary{positioning}
\usepackage{caption}
\usepackage{subcaption}
\usepackage{booktabs}
\usepackage{multirow}
\usepackage{multicol}
\usepackage{soul}
\usepackage{cite}
\usepackage[normalem]{ulem}
\usepackage{xcolor}
\usepackage{mathtools}
\usepackage{stfloats}

\newcommand{\dd}{\text{d}}
\let\labelindent\relax 
\usepackage{enumitem}
\usepackage{tabularx}
\usepackage{courier}
\usepackage{algorithm}
\usepackage{algorithmic}
\usepackage{makecell}

\usepackage{pifont}
\usepackage{url}

\newcommand{\todocite}[1]{\textcolor{red}{[TODO(cite)]}}

\allowdisplaybreaks

\newcommand{\norm}[1]{\lVert #1 \rVert}

\newtheorem{remark}{Remark}

\newcommand{\xinit}{\boldsymbol{x}_{\text{init}}}

\title{\LARGE 
\textbf{
DiffTune-MPC: Closed-Loop Learning for Model Predictive Control
}
}

\author{Ran Tao$^\dagger$, Sheng Cheng$^\dagger$, Xiaofeng Wang, Shenlong Wang, Naira Hovakimyan
\thanks{*This work is supported by NASA cooperative agreement (80NSSC22M0070), NASA ULI (80NSSC22M0070), AFOSR (FA9550-21-1-0411), NSF-AoF Robust Intelligence (2133656), and NSF SLES (2331878, 2331879). 
}
\thanks{$^\dagger$These authors contributed equally to this work. \newline
R. Tao, S. Cheng, S. Wang, and N. Hovakimyan are with the University of Illinois Urbana-Champaign, USA. ({\tt\small email: \{rant3, chengs, shenlong, nhovakim\}@illinois.edu})
\newline X. Wang is with the University of South Carolina, USA. ({\tt\small email: wangxi@cec.sc.edu})}
}

\begin{document}
\maketitle
\thispagestyle{plain}
\pagestyle{plain}

\begin{abstract}
Model predictive control (MPC) has been applied to many platforms in robotics and autonomous systems for its capability to predict a system's future behavior while incorporating constraints that a system may have. 
To enhance the performance of a system with an MPC controller, one can manually tune the MPC's cost function.
However, it can be challenging due to the possibly high dimension of the parameter space as well as the potential difference between the open-loop cost function in MPC and the overall closed-loop performance metric function. This paper presents DiffTune-MPC, a novel learning method, to learn the cost function of an MPC in a closed-loop manner. 
The proposed framework is compatible with the scenario where the time interval for performance evaluation and MPC's planning horizon have different lengths.
We show the auxiliary problem whose solution admits the analytical gradients of MPC and discuss its variations in different MPC
settings, including nonlinear MPCs that are solved using sequential quadratic programming. Simulation results demonstrate the learning capability of DiffTune-MPC and the generalization capability of the learned MPC parameters.
\end{abstract}

\begin{center}
    SUPPLEMENTARY INFORMATION
\end{center}
Code: https://github.com/RonaldTao/DiffTune-MPC\\


\section{Introduction}
\label{sec:intro}

With the advancement of computational power in embedded systems and the development of general-purpose solvers (e.g., acados, ACADO, CasADi),
Model Predictive Control (MPC) is being applied to more and more platforms in robotics and autonomous systems.
Effective implementation of MPC requires a well-designed cost function, a dynamical model with appropriate fidelity, realistic constraints accounting for the physical limits of the system, and sufficient computation power for real-time solutions of the MPC. 

When an MPC is applied for different tasks, one may need to design the cost function for each task to suit its individual characteristics (e.g., trajectory tracking tasks for hover or aggressive maneuvers for quadrotors). A common approach is to tune the parameters in a cost function with a given structure,
e.g., with weight matrices $Q$ and $R$ following the convention in a linear quadratic regulator (LQR). The challenges in manually tuning the $Q$ and $R$ matrices come from the following two aspects:

\noindent 1. Dimension: In the simplest case where $Q$ and $R$ are reduced to diagonal matrices, the parameter space's dimension equals the sum of the state's and control's dimensions, making manual tuning a challenging task, especially when the state is high-dimensional ($\geq$ 12 for 3D rigid-body dynamics).

\noindent 2. Difference between the cost function for MPC and loss function for MPC's performance: The cost function in MPC reflects the predictive cost in an \textit{open-loop} fashion: at each sample time, an optimal control problem is solved for a sequence of control actions into a future horizon, where all -- except for the first control actions -- are open-loop and not applied for controlling the system. On the contrary, the loss function is for evaluating the performance of the MPC as a function of the \textit{closed-loop} states (and control actions), e.g., tracking error between the desired trajectory and the closed-loop trajectory. In practice, the relationship between the cost function parameters and the performance judged by the loss function is complex and nonlinear. Furthermore, the loss function can be defined over a long horizon specified by an arbitrary task, whereas the cost in MPC is defined over a shorter horizon to enable real-time~solutions.

To address the challenges in the manual tuning of MPC's cost function, we propose to learn the cost function's parameters for a given task using DiffTune~\cite{cheng2022difftune}, an auto-tuning method for \textit{closed-loop} control systems.
DiffTune formulates the auto-tuning problem as a parameter optimization problem that can handle different horizons in the loss function (for performance) and cost function (for control). It has been previously demonstrated on explicitly differentiable controllers whose Jacobian can be conveniently obtained by autodifferentiation. In this work, we extend DiffTune to implicitly differentiable MPC controllers.
The analytical gradients of the MPC policy are obtained by the implicit differentiation proposed in~\cite{amos2018differentiable}, which is based on the implicit function theorem and Karush–Kuhn–Tucker (KKT) conditions. We first show how to differentiate a linear MPC problem (with quadratic cost, linear dynamics, and linear inequality constraints) by solving an auxiliary linear MPC. Then, we extend it to a nonlinear MPC and show how to differentiate it utilizing the auxiliary MPC. We validate our approach in simulations, where we show the efficacy of DiffTune for MPC subject to both linear and nonlinear dynamical systems. We also validated the generalization capability and applicability to real-world scenarios in a high-fidelity simulator.

We summarize our contributions as follows:

\begin{itemize}
    \item We extend the application of DiffTune~\cite{cheng2022difftune} from auto-tuning explicitly differentiable controllers to MPC that is implicitly differentiable.
    \item We propose a novel and flexible closed-loop MPC learning scheme using differentiable programming for the general scenario where the horizon for closed-loop performance evaluation is longer than an MPC's open-loop planning horizon. This is a significant improvement compared to existing open-loop MPC learning schemes~\cite{amos2018differentiable,jin2021safe} that apply differentiable programming, where the horizon of the loss function cannot be longer than the MPC's horizon because the loss function therein is defined with respect to the open-loop optimal solution.
    \item Our differentiation of the MPC is based on a general formulation of a linear MPC that contains linear-inequality constraints of state and control, which covers a broader range of applications than the boxed-control-constraints only formulation in~\cite{amos2018differentiable}.
\end{itemize}

The remainder of the paper is organized as follows: Sections~\ref{sec: related work} and~\ref{sec: tech background} review related work and background, respectively.
Section ~\ref{sec:prob} describes the problem formulation, and section~\ref{sec: method} describes our method for differentiating MPC based on a general linear MPC formulation. Section~\ref{sec: simulation} shows the simulation results on a unicycle model with linear inequality constraints, a quadrotor model, and a 1D double integrator model. Section~\ref{sec:experiment} presents the results on a quadrotor model using a high-fidelity simulator to show the generalization capability.
Section~\ref{sec: conclusion} concludes the paper.
\section{Related Work}\label{sec: related work}
\textbf{Auto-tuning}, in general, can be formulated as a parameter optimization problem that looks for an optimal parameter to minimize a loss function subject to a given controller. 
{It can be considered as a special case of learning for control, where the structure of the controller comes from physics (e.g., by Lyapunov or conventional design) rather than a generic neural network.}
If the system model is available, one can leverage the model information to infer optimal parameter choice, e.g., using gradient descent~\cite{parwana2021recursive,cheng2022difftune,cheng2023difftuneplus,kumar2021diffloop}. If the model is unavailable, then one needs to query the system's performance from a chosen set of parameters to determine the next choice of the parameter. For example, the authors of~\cite{berkenkamp2016safe,duivenvoorden2017constrained} use Bayesian optimization to search for an optimal parameter by treating the mapping from parameter to loss as a Gaussian process. The parameters are iteratively chosen as the minimizer with the maximum likelihood. In~\cite{song2022policy}, a policy search method is proposed to select high-level decision variables or parameters for MPC automatically. The authors of~\cite{loquercio2022autotune} use the Metropolis-Hastings sampling to generate random parameters and selectively keep one parameter at each trial to continue the sampling procedure.
A data-driven approach is proposed in~\cite{edwards2021automatic}, which consists of a system identification module, an objective specification module, and a feedback control synthesis module, where the controller is optimized end-to-end using Bayesian optimization. This method is applied to control a challenging soft robot platform~\cite{null2023automatically}.

\textbf{Learning for MPC} has been a trending topic in recent years owing to the hardware and software advancement in deploying and training MPC-related algorithms. The concept of differentiable MPC is initially discussed in~\cite{amos2018differentiable} and later extended to general optimal control problems~\cite{jin2020pontryagin,jin2021safe} from the perspective of Pontryagin's maximum principle and robust MPC~\cite{oshin2023differentiable}. The structural similarity between optimal control and reinforcement learning (RL) has made differentiable MPC a bridge to connect the two worlds~\cite{zanon2020safe}. 
In~\cite{wiedemann2023training}, a receding-horizon implementation of analytical policy gradient is proposed with a curriculum learning scheme to improve training stability for offline training of a controller.
The authors of~\cite{romero2023actor} propose to use the actor-critic method to learn the parameters of an MPC policy, which combines the long-term learning capability of the actor-critic RL and the short-term control capability by MPC. One particular direction for learning-based MPC uses learning to enhance the model's quality for better prediction performance (and thus control performance). Various approaches~\cite{torrente2021data,chee2022knode,chee2023enhancing,jiahao2023online,saviolo2022physics} 
have been proposed to learn the residual dynamics and show promising performance for quadrotor control.

In contrast to previous works~\cite{amos2018differentiable,jin2020pontryagin} that apply differentiable programming to parameter learning of optimal control (OC) problems in an open-loop fashion, our approach tackles parameter learning of an MPC problem in a \textit{closed-loop} fashion. \footnote{Note that the difference between OC and MPC is that MPC refers to the practice of solving an OC but only applies the first optimal control action back to the system. }
In an open-loop setting,~\cite{amos2018differentiable,jin2020pontryagin} formulate the loss for training as a function of the optimal states and control actions that are obtained from solving the OC problem \textit{once}. This design restricts the loss function's horizon such that it cannot be longer than the MPC's horizon, limiting suitable applications (e.g., imitation learning) because the loss's horizon for an arbitrary task is usually much longer than the MPC's planning horizon. 
In a closed-loop setting, our loss function is defined over a closed-loop system's state and control actions, which allows for a broad set of learning applications where the loss's horizon is way longer than the MPC's horizon (which cannot be handled by open-loop learning~\cite{amos2018differentiable,jin2020pontryagin}). The closed-loop learning of an MPC has been explored using probabilistic approaches like policy search~\cite{song2022policy} and Bayesian optimization~\cite{edwards2021automatic}. These approaches do not employ first-order information of the MPC and, therefore, are less efficient than those that do (e.g., this work).

In addition to the different problem formulation and application scenarios discussed above, our approach has novelties in terms of differentiating the OC problem. The differentiable OC presented in previous work~\cite{amos2018differentiable} is based on an LQR with boxed constraints for control actions only. We extend it to a more general case of LQR with linear inequality constraints, which are capable of handling state constraints for safety requirements and likely coupled control(-state) constraints. 
Towards the incorporation of constraints, our OC problem's formulation is similar to~\cite{jin2021safe} albeit with different solution and differentiation techniques.
The results in \cite{jin2021safe} augment the constraints (excluding the dynamics) to the cost function using barrier functions and thus resort to  Pontryagin's differentiable programming~\cite{jin2020pontryagin} to solve the OC as well as find the analytical gradients. In this procedure, one needs to manually select the barriers' multipliers for a tradeoff between constraint tightness and cost minimization. In addition, augmenting the constraints to soft constraints makes the resulting solution only an approximate one for the original problem. In contrast, our method is based on the KKT condition of an MPC to satisfy the hard constraints. The KKT-based formulation is also general enough to use off-the-shelf MPC solvers (e.g., acados) to obtain both the MPC solution and the analytical gradient.

\section{Background} \label{sec: tech background}

The following notation will be used throughout the paper: 
For a matrix $A$, we use $\underline{A}$ to denote its vectorized form (by column). We use $[\boldsymbol{v}]_j$ to denote the $j$-th element of vector $\boldsymbol{v}$. Similarly, $[A]_{i,:}$, $[A]_{:,j}$ and $[A]_{i,j}$  denote the $i$-th row, $j$-th column, and element on $i$-th row and $j$-th column of the matrix $A$, respectively. The matrix $I_n$ denotes an $n$-dimensional identity matrix. The vector $\mathbf{1}_n$ denotes an $n$-dimensional vector with all elements being 1. The vector $\boldsymbol{e}_i^n$ denotes an $n$-dimensional unit vector with the $i$-th element being 1.

We briefly review controller auto-tuning and DiffTune in the sequel.
Consider a discrete-time dynamical system
\vspace{-1mm}
\begin{equation}\label{eq: dynamics}
    \boldsymbol{x}_{k+1} = f(\boldsymbol{x}_k,\boldsymbol{u}_k), 
    \vspace{-1mm}
\end{equation}

where $\boldsymbol{x}_k \in \mathbb{R}^n$ and $\boldsymbol{u}_k \in \mathbb{R}^m$ are the state and control, respectively, and the initial state $\boldsymbol{x}_0$ is known. The control is generated by a feedback controller that tracks a desired state $\bar{\boldsymbol{x}}_{k} \in \mathbb{R}^n$ such that\vspace{-1mm}
\begin{equation}
    \boldsymbol{u}_k = h(\boldsymbol{x}_k,\bar{\boldsymbol{x}}_{k},\boldsymbol{\theta}), \label{eq: feedback controller}
    \vspace{-1mm}
\end{equation}
where $\boldsymbol{\theta} \in \Theta$ denotes the controller's parameters, and $\Theta$ represents a feasible set of parameters that can be analytically or empirically determined for a system's stability.
We assume that the state $\boldsymbol{x}_k$ can be measured directly or, if not, an appropriate state estimator is used. Furthermore, we assume that the dynamics \eqref{eq: dynamics} and controller \eqref{eq: feedback controller} are differentiable, i.e., the Jacobians $\nabla_{\boldsymbol{x}}f$, $\nabla_{\boldsymbol{u}} f$, $\nabla_{\boldsymbol{x}}h$, and $\nabla_{\boldsymbol{\theta}} h$ exist, which widely applies to general systems.

The controller learning/auto-tuning task adjusts $\boldsymbol{\theta}$ to minimize an evaluation criterion, denoted by a loss function $L(\cdot)$, which is a function of the desired states, actual states, and control actions over a time interval of length $N$. An illustrative example is the quadratic loss of
the tracking error and control-effort penalty, where $L(\boldsymbol{x}_{1:N},\bar{\boldsymbol{x}}_{1:N},\boldsymbol{u}_{0:N-1};\boldsymbol{\theta}) = \sum_{k=1}^{N} \norm{\boldsymbol{x}_k -\bar{\boldsymbol{x}}_{k}}^2 + \sum_{k=0}^{N-1}\lambda \norm{\boldsymbol{u}_k}^2$ 
with $\lambda>0$ being the penalty coefficient. We will use the short-hand notation $L(\boldsymbol{\theta})$ for conciseness in the rest of the paper. Note that the usage of DiffTune is not limited to the above example of a tracking problem. As long as a loss function can be specified and is differentiable, we are able to apply DiffTune to learn the parameters and minimize the loss.

To summarize, we formulate the controller learning/auto-tuning problem as the following parameter optimization problem
\begin{equation}\label{prob: controller tuning as a parameter optimization}
    \begin{aligned} 
    & \underset{\boldsymbol{\theta} \in \Theta}{\text{minimize}} && L(\boldsymbol{\theta})\\
    & \text{subject to} && \boldsymbol{x}_{k+1} = f(\boldsymbol{x}_k,\boldsymbol{u}_k), \ {x}_0 \text{ given},\\
    & && \boldsymbol{u}_k \! =\! h(\boldsymbol{x}_k,\bar{\boldsymbol{x}}_{k},\boldsymbol{\theta}),\!\ k \in \{0,1,\dots,N-1\}.
    \end{aligned} \nonumber
    \tag{P}
\end{equation}
We solve the above problem by DiffTune, which uses a projected gradient descent algorithm~\cite{parikh2014proximal} to update the parameters iteratively:
\begin{equation}\label{eq: projected gradient descent to update the parameters}
    \boldsymbol{\theta} \leftarrow \mathcal{P}_{\Theta} ( \boldsymbol{\theta} - \alpha \nabla_{\boldsymbol{\theta}}L).
\end{equation}
To obtain the gradient $\nabla_{\boldsymbol{\theta}} L $, we start with the chain rule: 
\begin{equation}
    \nabla_{\boldsymbol{\theta}}L =\sum_{k=1}^N \frac{\partial L}{\partial \boldsymbol{x}_k} \frac{\partial \boldsymbol{x}_k}{\partial \boldsymbol{\theta}} + \sum_{k=0}^{N-1} \frac{\partial L}{\partial \boldsymbol{u}_k} \frac{\partial \boldsymbol{u}_k}{\partial \boldsymbol{\theta}}.\label{eq: decomposition of the target derivative to accepting sensitivity propagation}
\end{equation}
The gradients $\partial L / \partial \boldsymbol{x}_k$ and $\partial L / \partial \boldsymbol{u}_k$ can be determined once $L$ is chosen, and we use sensitivity propagation to obtain the sensitivity states $\partial \boldsymbol{x}_k / \partial \boldsymbol{\theta}$ and $\partial \boldsymbol{u}_k / \partial \boldsymbol{\theta}$ by:
\begin{subequations}\label{eq: sensitivity propagation}
\begin{align}
    \frac{\partial \boldsymbol{x}_{k+1}}{\partial \boldsymbol{\theta}} = & (\nabla_{\boldsymbol{x}_k} f + \nabla_{\boldsymbol{u}_k} f \nabla_{\boldsymbol{x}_k} h) \frac{\partial \boldsymbol{x}_k}{\partial \boldsymbol{\theta}} + \nabla_{\boldsymbol{u}_k} f \nabla_{\boldsymbol{\theta}} h, \label{eq: iterative Jacobian of state wrt parameter} \\
    \frac{\partial \boldsymbol{u}_k}{\partial \boldsymbol{\theta}} = & \nabla_{\boldsymbol{x}_k} h \frac{\partial \boldsymbol{x}_k}{\partial \boldsymbol{\theta}} + \nabla_{\boldsymbol{\theta}} h, \label{eq: iterative Jacobian of control wrt parameter}
\end{align}
\end{subequations}
with $\partial \boldsymbol{x}_0/\partial \boldsymbol{\theta} = 0$.

Here, the notations $\nabla_{\boldsymbol{x}_k} f$, $\nabla_{\boldsymbol{u}_k} f$, $\nabla_{\boldsymbol{x}_k} h$, and $\nabla_{\boldsymbol{\theta}} h$ mean the Jacobians $\partial f / \partial \boldsymbol{x}$, $\partial f / \partial \boldsymbol{u}$, $\partial h / \partial \boldsymbol{x}$, and $\partial h / \partial \boldsymbol{\theta}$ is evaluated at the state $\boldsymbol{x}_k$ and control $\boldsymbol{u}_k$. {We allow the state $\boldsymbol{x}_k$ to come from sensor measurements or state estimation, which permits data collected from a real system or simulated noisy system (beyond the data collected by rolling out the dynamics~\eqref{eq: dynamics}).}

\section{Problem Formulation}\label{sec:prob}
Model predictive control determines a sequence of control actions at every step that minimizes a cost criterion subject to system dynamics and other constraints on the states and control inputs. The problem is repeatedly solved when new state measurement becomes available, and the first optimal control action is implemented for the closed-loop control.
In general, an MPC problem has the following form: 
\begin{align}\label{prob: general MPC formulation}
    & \underset{\boldsymbol{x}_{1:T},\boldsymbol{u}_{1:T}}{\text{minimize}} && \sum_{t=1}^T \mathcal{C}_t(\boldsymbol{x}_t,\boldsymbol{u}_t) \nonumber \\
    & \text{subject to} && \boldsymbol{x}_{t+1} = f(\boldsymbol{x}_t,\boldsymbol{u}_t), \ \boldsymbol{x}_1 = \boldsymbol{x}_{\text{init}}, \nonumber\\
    & && \boldsymbol{x}_t \in \mathcal{X}_t, \ \boldsymbol{u}_t \in \mathcal{U}_t, \ t \in \{0,1,\dots,T-1\}, \nonumber\\
    & && \boldsymbol{x}_T \in \mathcal{X}_T,
    \tag{MPC}
\end{align}
where $\mathcal{X}_t$ and $\mathcal{U}_t$ stand for feasible sets of states and control actions at time $t$, and $\mathcal{X}_T$ denotes the terminal set of states.
The simplest case of an MPC would be an LQR, i.e., $\mathcal{C}_t(\boldsymbol{x}_t,\boldsymbol{u}_t) = \frac{1}{2} (\boldsymbol{x}_t-\bar{\boldsymbol{x}}_t)^\top Q_t (\boldsymbol{x}_t-\bar{\boldsymbol{x}}_t) + \frac{1}{2}(\boldsymbol{u}_t-\bar{\boldsymbol{u}}_t)^\top R_t (\boldsymbol{u}_t-\bar{\boldsymbol{u}}_t)$ and $f(\boldsymbol{x}_t,\boldsymbol{u}_t) = A_t \boldsymbol{x}_t + B_t \boldsymbol{u}_t$ in MPC, where $Q_t, R_t, A_t, B_t$ are matrices with appropriate dimensions.  No other constraints are enforced in LQR.

\begin{remark}
    Note that we distinguish the time indices used in DiffTune ($k$ and $N$ for closed-loop systems) and in MPC ($t$ and $T$ for open-loop MPC solution). For the closed-loop system, at time $k$, the instantaneous state $\boldsymbol{x}_k$ is set to $\xinit$ in \eqref{prob: general MPC formulation}. Once the \eqref{prob: general MPC formulation} is solved, its initial optimal control action $\boldsymbol{u}_1^\star$ is set as $\boldsymbol{u}_k$ for the closed-loop system. Furthermore, the horizon $N$ of the performance evaluation in DiffTune is generally much larger than the horizon $T$ in \eqref{prob: general MPC formulation}.
\end{remark}

The goal of this paper is to apply DiffTune to solve the problem \eqref{prob: controller tuning as a parameter optimization} when the controller $h(\cdot)$ is an MPC. More specifically, we want to learn the cost functions of the MPC controller to minimize the loss function defined over the entire closed-loop trajectory.
Since MPC is an implicitly differentiable controller, the Jacobians $\partial h / \partial \boldsymbol{\theta}$ and $\partial h / \partial \boldsymbol{x}_k$ cannot be calculated directly like a simple feedback control law $\boldsymbol{u} = K \boldsymbol{x}$. Thus, we will present our method to efficiently compute the target Jacobians in the following section.

\section{The DiffTune-MPC Method} \label{sec: method}

We learn an MPC controller following a given loss function by auto-tuning the parameter $\boldsymbol{\theta}$ in the cost function (e.g., $Q$ and $R$ for quadratic cost) using DiffTune. 
Following the sensitivity propagation in \eqref{eq: sensitivity propagation}, at each time $k$, we need to know the Jacobians $\partial h / \partial \boldsymbol{\theta}$ and $\partial h / \partial \boldsymbol{x}_k$, which are the sensitivity of the controller to the parameter $\boldsymbol{\theta}$ and instantaneous state $\boldsymbol{x}_k$. 
In the setting of using MPC as a controller $h$ at step $k$, we have $\partial h / \partial \boldsymbol{\theta} = \partial \boldsymbol{u}_1^\star / \partial \boldsymbol{\theta}$ and $\partial h / \partial \boldsymbol{x}_k = \partial \boldsymbol{u}_1^\star / \partial \xinit$, where $\boldsymbol{u}_1^\star$ is the first optimal control action obtained by solving problem~\eqref{prob: general MPC formulation} with the initial state $\xinit = \boldsymbol{x}_k$. 
We will leverage the KKT conditions and implicit differentiation to obtain the gradients as illustrated in~\cite{amos2018differentiable}. We start by differentiating a linear MPC problem and discuss the differentiation of a nonlinear MPC at the end. Considering the page limit, we only present the main results in this section. The complete derivation and analysis can be found in the Appendix.


\subsection{Differentiating a Linear MPC}\label{subsec: diff linear MPC}
By linear MPC, we refer to the following problem
\begin{align}
    & \underset{\boldsymbol{\tau}_{1:T}}{\text{minimize}} && \sum_{t=1}^T \frac{1}{2}\boldsymbol{\tau}_t^\top C_t \boldsymbol{\tau}_t + \boldsymbol{c}_t^\top \boldsymbol{\tau}_t \nonumber \\
    & \text{subject to} && \boldsymbol{x}_{t+1} = F_t \boldsymbol{\tau}_t + \boldsymbol{f}_t, \ \boldsymbol{x}_{1} = \xinit, \nonumber \\
    & && G_t \boldsymbol{\tau}_t \leq \boldsymbol{l}_t,
    \nonumber
    \tag{LMPC}
    \label{prob: LQR with linear inequalities}
\end{align}
where $\boldsymbol{\tau}_t = [\begin{smallmatrix}
    \boldsymbol{x}_t \\ \boldsymbol{u}_t
\end{smallmatrix}] \in \mathbb{R}^{n+m}$ is the composite state-control to be optimized; $C_t\in \mathbb{R}^{(n+m)\times(n+m)}$ and $\boldsymbol{c}_t \in \mathbb{R}^{n+m}$ are the cost function parameters; $F_t = [   A_t \ B_t] \in \mathbb{R}^{n \times (n+m)}$; $\boldsymbol{f}_t \in \mathbb{R}^n$ stands for residuals in the dynamics; and $G_t \in \mathbb{R}^{g \times (n+m)}$ and $\boldsymbol{l}_t \in \mathbb{R}^g$ stand for $g$ number of inequalities 
on the state and control actions. 
Although a more general form of inequality constraints for $\boldsymbol{\tau}_t$ of all $t \in \{1,2,\dots,T\}$ can be incorporated, we stick to the form displayed in \eqref{prob: LQR with linear inequalities} where the linear inequality constraints are separable by $t$.
The commonly used boxed constraint on the control action is a special case of the form used in $G_t \boldsymbol{\tau}_t \leq h_t$, where $G_t = [\begin{smallmatrix}
    0 & I_m \\ 0 & -I_m
\end{smallmatrix}]$ and $h_t = [\begin{smallmatrix}
    u_{\max} \mathbf{1}_m \\ -u_{\min} \mathbf{1}_m
\end{smallmatrix}] $ define the bounds of each element of the control to be bounded by $u_{\max}$ and $u_{\min}$, respectively.

Problem \eqref{prob: LQR with linear inequalities} is a quadratic program (QP) and can be solved efficiently with off-the-shelf solvers. Once we get the optimal solution $\{\boldsymbol{\tau}_t^\star\}_{t = 1:T}$, we can proceed to solve the following auxiliary problem for the gradients $\partial \boldsymbol{u}_1^\star / \partial [C_t]_{i,j}$, $\partial \boldsymbol{u}_1^\star / \partial [\boldsymbol{c}_t]_{i}$, and $\partial \boldsymbol{u}_1^\star / \partial [\xinit]_{i}$: 
\begin{align}
    & \underset{\tilde{\boldsymbol{\tau}}_{1:T}}{\text{minimize}} && \sum_{t=1}^T \frac{1}{2}\tilde{\boldsymbol{\tau}}_t^\top C_t \tilde{\boldsymbol{\tau}}_t + \tilde{\boldsymbol{c}}_t^\top \tilde{\boldsymbol{\tau}}_t \nonumber \\
    & \text{subject to} && \tilde{\boldsymbol{x}}_{t+1} = F_t \tilde{\boldsymbol{\tau}}_t , \ \tilde{\boldsymbol{x}}_1 = \tilde{\boldsymbol{x}}_{\text{init}}, \nonumber \\
    & && [G_t]_{\mathcal{I}_{t,\text{act}}} \tilde{\boldsymbol{\tau}}_t = 0,
    \nonumber
    \tag{LMPC-Grad}
    \label{prob: alt problem to solve for the analytical gradient}
\end{align}
where $\boldsymbol{\tilde{\tau}}_t = [\begin{smallmatrix}
    \boldsymbol{\tilde{x}}_t \\ \boldsymbol{\tilde{u}}_t
\end{smallmatrix}] \in \mathbb{R}^{n+m}$ is the composite auxiliary state-control to be optimized, $[G_t]_{\mathcal{I}_{t,\text{act}}}$ denote the rows of $G_t$ where the inequality $G_t \boldsymbol{\tau}_t \leq \boldsymbol{l}_t$ is active at the optimal solution $\boldsymbol{\tau}_t^\star$, i.e., $[G_t]_{\mathcal{I}_{t,\text{act}}} \boldsymbol{\tau}_t^\star = [\boldsymbol{l}_t]_{\mathcal{I}_{t,\text{act}}}$.
Denote the first optimal control action of~\eqref{prob: alt problem to solve for the analytical gradient} by $\tilde{\boldsymbol{u}}_1^\star$.
The analytical gradients for \eqref{prob: LQR with linear inequalities} can be obtained by computing $\tilde{\boldsymbol{u}}_1^\star$ from \eqref{prob: alt problem to solve for the analytical gradient} with the coefficients $\tilde{\boldsymbol{c}}_t$ and $\tilde{\boldsymbol{x}}_{\text{init}}$ listed in Table \ref{tb: table for gradient computation with C c parametrization}. 
For details in the derivation of \eqref{prob: alt problem to solve for the analytical gradient} for computing the analytical gradient, see Appendix~C. 
Once we obtain the target gradients, we can leverage \eqref{eq: decomposition of the target derivative to accepting sensitivity propagation} and \eqref{eq: sensitivity propagation} to obtain $\nabla_{\theta}L$ and update the parameter using \eqref{eq: projected gradient descent to update the parameters}.

\begin{remark}
The high-level idea of obtaining the analytical gradients of \eqref{prob: LQR with linear inequalities} by solving \eqref{prob: alt problem to solve for the analytical gradient} 
    is summarized as follows: Problem \eqref{prob: LQR with linear inequalities} is solved by treating it as a QP, where the optimal solution satisfies a linear equation~\eqref{eq: KKT factorization for LQR-LI} following the KKT condition (including the dynamics, costate dynamics, boundary conditions, and constraints). One can take matrix differentials~\cite{amos2017optnet} of this linear equation, which governs the relation among the infinitesimal changes of the decision variables (state, control, Lagrange multipliers) and parameters (such as cost matrix $C_t$). Meanwhile, the resulting equation of matrix differentials~\eqref{eq: KKT factorization for differentials of LQR-LI} leads to the governing linear equation of the gradients of \eqref{prob: LQR with linear inequalities}, which coincides with the KKT condition of the auxiliary problem~\eqref{prob: alt problem to solve for the analytical gradient}. Hence, one can match the terms and coefficients in the original KKT condition with those in the one derived from the matrix differentials, which allows us to obtain the target gradients listed in the first column of Table~\ref{tb: table for gradient computation with C c parametrization} by solving \eqref{prob: alt problem to solve for the analytical gradient} with associated terms listed in the last two columns of Table~\ref{tb: table for gradient computation with C c parametrization}.
\end{remark}

\setlength{\tabcolsep}{6pt} 
\renewcommand{\arraystretch}{1} 
  \captionsetup{
	skip=5pt, position = bottom}
\begin{table}[h]
	\centering
	\small
	
	\vspace{-0.2cm}
	\captionsetup{font=small}
	\caption{To get target gradients of \eqref{prob: LQR with linear inequalities} with $(C_t,\boldsymbol{c}_t)$-parametrization in the left column, one can solve \eqref{prob: alt problem to solve for the analytical gradient} with coefficients $\tilde{\boldsymbol{c}}_t$ and $\tilde{\boldsymbol{x}}_{\text{init}}$ specified in the right column. The solution $\tilde{\boldsymbol{u}}_1^\star$ of \eqref{prob: alt problem to solve for the analytical gradient} equals the gradient in the left column. Note that the values in the column of $\Tilde{\boldsymbol{c}}_t$ are specifically for time step $t$ from the target parameters $C_t$ and $\boldsymbol{c}_t$, and for all other time steps $t'\neq t$, we have $\tilde{\boldsymbol{c}}_{t'}=0$.} 
\begin{tabular}{c|cc}
		\toprule[1pt]
		Target gradients in & \multicolumn{2}{c}{Coefficients in~\eqref{prob: alt problem to solve for the analytical gradient} }
		\\
		\eqref{prob: LQR with linear inequalities} & $\tilde{\boldsymbol{c}}_t$ & $\tilde{\boldsymbol{x}}_{\text{init}}$ \\
        \midrule
        $\frac{\partial \boldsymbol{u}_1^\star}{\partial [{C}_t]_{i,j}}$ &  $[\boldsymbol{\tau}_t^\star]_j \boldsymbol{e}_i^{n+m}$ & $0$ \\
        $\frac{\partial \boldsymbol{u}_1^\star}{\partial [\boldsymbol{c}_t]_{i}}$ &  $\boldsymbol{e}_i^{n+m}
        $ & $0$ \\
        $\frac{\partial \boldsymbol{u}_1^\star}{ \partial [\xinit]_{i}}$ & $0$ & $\boldsymbol{e}_i^n$ \\
       	\bottomrule[1pt]
	\end{tabular}\label{tb: table for gradient computation with C c parametrization}
\end{table}

\begin{remark}\label{remark: computation time}
{The analytical gradient obtained by solving~\eqref{prob: alt problem to solve for the analytical gradient} is computationally cheaper than a numerical gradient obtained by finite difference. 
Since \eqref{prob: alt problem to solve for the analytical gradient} is a QP with \textit{equality constraints only} (solving such a problem reduces to solving a linear system of equations~\cite{boggs1995sequential}), \eqref{prob: alt problem to solve for the analytical gradient} takes less time to solve than the original problem~\eqref{prob: LQR with linear inequalities} which contains inequality constraints.
If one wants to get the analytical gradient of $\boldsymbol{u}_1^\star$ to $M$ scalar parameters in \eqref{prob: LQR with linear inequalities}, then one needs to solve~\eqref{prob: alt problem to solve for the analytical gradient}  $M$ times. However, if one wants to get the numerical gradient, then one has to solve at least $M$ times  \eqref{prob: LQR with linear inequalities} with single-direction perturbations for $M$  parameters and perform finite difference. Besides, the quality of the numerical gradient depends on the size of the perturbation, determining which can be an individual problem of study.
}

\end{remark}

In the problem setting of DiffTune, when differentiating~\eqref{prob: LQR with linear inequalities} through solving~\eqref{prob: alt problem to solve for the analytical gradient}, we only need $\tilde{\boldsymbol{\tau}}_1$ (more specifically $\tilde{\boldsymbol{u}}_1$). Hence, if $\boldsymbol{u}_1^\star$ is active at some constraints (i.e., $\mathcal{I}_{1,\text{act}} \neq \emptyset$), then the gradient of \eqref{prob: LQR with linear inequalities} $\partial \boldsymbol{u}_1^\star / \partial \boldsymbol{\theta}$ (for $\boldsymbol{\theta}$ being an arbitrary scalar element of $C_t$ or $\boldsymbol{c}_t$) satisfies
\begin{equation}\label{eq: linear relationship between derivatives subject to the linear constraint}
    [G_1]_{\mathcal{I}_{1,\text{act}}} \tilde{\boldsymbol{\tau}}_1  =  [G_1]_{\mathcal{I}_{1,\text{act}}} \left[ \begin{smallmatrix}
        \frac{\partial \boldsymbol{x}_{1}^\star}{\partial \boldsymbol{\theta}} \\ \frac{\partial \boldsymbol{u}_1^\star}{\partial \boldsymbol{\theta}}
    \end{smallmatrix} \right]  = 0.
\end{equation}
Consider a simple example of a scalar control $\boldsymbol{u}_1$ to be upper bounded by $\boldsymbol{u}_{\max}$, where $G_t = [0 \ 1]$ and $\boldsymbol{l}_t = \boldsymbol{u}_{\max}$ for all $t$. If $\boldsymbol{u}_1^\star = \boldsymbol{u}_{\max}$, then $\partial \boldsymbol{u}_1^\star / \partial \boldsymbol{\theta} = 0$, indicating the inefficacy of manipulating $\boldsymbol{u}_1^\star$ by tweaking $\boldsymbol{\theta}$ since $\boldsymbol{u}_1^\star$ will remain $\boldsymbol{u}_{\max}$ when $\boldsymbol{\theta}$ is perturbed locally. 
A detailed discussion on active constraints and corresponding gradients is available in Appendix~F.
Simulation results showing the degraded efficiency in learning when control saturation happens are included in Appendix~E.

\subsection{Linear MPC with interpretable quadratic cost function}\label{subsec: quadratic cost linear MPC}
Although the $(C_t,\boldsymbol{c}_t)$-parametrization is general enough, its physical meaning is not as clear as the conventional $(Q_t,R_t)$-parametrization. By $(Q_t,R_t)$-parametrization, we consider the cost function of the form $\mathcal{C}_t(\boldsymbol{x}_t,\boldsymbol{u}_t) = \frac{1}{2} (\boldsymbol{x}_t-\bar{\boldsymbol{x}}_t)^\top Q_t (\boldsymbol{x}_t-\bar{\boldsymbol{x}}_t) + \frac{1}{2} (\boldsymbol{u}_t-\bar{\boldsymbol{u}}_t)^\top R_t (\boldsymbol{u}_t-\bar{\boldsymbol{u}}_t)$, where $\bar{\boldsymbol{x}}_t$ and $\bar{\boldsymbol{u}}_t$ are the reference state and control to be tracked. In this case, the parameters to be learned are $Q_t\in \mathbb{R}^{n \times n}$ and $R_t\in \mathbb{R}^{m \times m}$, and this cost function associates with the one in \eqref{prob: LQR with linear inequalities} by $C_t = \text{diag}(Q_t,R_t)$ and $\boldsymbol{c}_t = -[\begin{smallmatrix} Q_t & \\ & R_t\end{smallmatrix}] \bar{\boldsymbol{\tau}_t}$ for $\bar{\boldsymbol{\tau}}_t = [\begin{smallmatrix}
    \bar{\boldsymbol{x}}_t \\ \bar{\boldsymbol{u}}_t
\end{smallmatrix}]$. Similar to Table~\ref{tb: table for gradient computation with C c parametrization}, if one wants to solve for $\partial \boldsymbol{u}_1^\star / \partial [Q_t]_{i,j}$ and $\partial \boldsymbol{u}_1^\star / \partial [R_t]_{i,j}$, then one can simply plug in $\tilde{\boldsymbol{c}}_t$ into problem \eqref{prob: alt problem to solve for the analytical gradient} as indicated in Table~\ref{tb: table for gradient computation with Q R parametrization} and solve \eqref{prob: alt problem to solve for the analytical gradient} for the target gradients.

\setlength{\tabcolsep}{6pt} 
\renewcommand{\arraystretch}{1} 
  \captionsetup{
	skip=5pt, position = bottom}
\begin{table}[b]
	\centering
	\small
	
	\vspace{-0.2cm}
	\captionsetup{font=small}
	\caption{To get target gradients of \eqref{prob: LQR with linear inequalities} with $(Q_t,R_t)$-parametrization in the left column, one can solve \eqref{prob: alt problem to solve for the analytical gradient} with coefficients $\tilde{\boldsymbol{c}}_t$ and $\tilde{\boldsymbol{x}}_{\text{init}}$ specified in the right column. The solution $\tilde{\boldsymbol{u}}_1^\star$ of \eqref{prob: alt problem to solve for the analytical gradient} equals the gradient of \eqref{prob: LQR with linear inequalities} in the left column.}
\begin{tabular}{c|cc}
		\toprule[1pt]
		Target gradients in & \multicolumn{2}{c}{Coefficients in~\eqref{prob: alt problem to solve for the analytical gradient} }
		\\
		\eqref{prob: LQR with linear inequalities} & $\tilde{\boldsymbol{c}}_t$ & $\tilde{\boldsymbol{x}}_{\text{init}}$ \\
        \midrule
        $\frac{\partial \boldsymbol{u}_1^\star}{\partial [{Q}_t]_{i,j}}$ &  $\begin{bmatrix}
            [\boldsymbol{x}_t^\star - \bar{\boldsymbol{x}}_t]_j \boldsymbol{e}_i^n\\0
        \end{bmatrix}$ & $0$ \\
        $\frac{\partial \boldsymbol{u}_1^\star}{\partial [{R}_t]_{i,j}}$ &  $\begin{bmatrix}
            0\\ [\boldsymbol{u}_t^\star - \bar{\boldsymbol{u}}_t]_j \boldsymbol{e}_i^m
        \end{bmatrix}$ & $0$ \\
       	\bottomrule[1pt]
	\end{tabular}\label{tb: table for gradient computation with Q R parametrization}
\end{table}

\subsection{Nonlinear MPC}
The linear MPC problem discussed above does not cover practical control problems in reality, where one can have general nonlinear dynamics, costs, and constraints. For such nonlinear MPC problems, we consider the following formulation:
\begin{align}\label{prob: nonlinear MPC formulation}
    & \underset{\boldsymbol{\tau}_{1:T}}{\text{minimize}} && \sum_{t=1}^T \mathcal{C}_{\boldsymbol{\theta},t}(\boldsymbol{\tau}_t) \nonumber \\
    & \text{subject to} && \boldsymbol{x}_{t+1} = f(\boldsymbol{\tau}_t), \ \boldsymbol{x}_{1} = \xinit, \nonumber\\
    & && g(\boldsymbol{\tau}_t) \le 0,
    \tag{NMPC}
\end{align}
where the nonlinear cost function $\mathcal{C}_{\boldsymbol{\theta},t}$ is parameterized by $\boldsymbol{\theta}$, $f$ is the nonlinear dynamics, and $g$ is the nonlinear inequality constraint function. 

One common way to solve this nonlinear optimization problem is to treat it as nonlinear programming and apply sequential quadratic programming (SQP) \cite{boggs1995sequential}, which iteratively generates and optimizes the following approximate problem:
\begin{align}
    & \underset{\boldsymbol{\tau}_{1:T}}{\text{minimize}} && \sum_{t=1}^T \frac{1}{2} \boldsymbol{\tau}_t^\top \nabla^2 \mathcal{C}_{\boldsymbol{\theta},t}^p \boldsymbol{\tau}_t + (\nabla \mathcal{C}_{\boldsymbol{\theta},t}^p - \nabla^2 \mathcal{C}_{\boldsymbol{\theta},t}^p \boldsymbol{\tau}_t^p)^\top \boldsymbol{\tau}_t \nonumber\\
    & \text{subject to } &&\boldsymbol{x}_{t+1} = \nabla f^p \boldsymbol{\tau}_t + f^p - \nabla f^p \boldsymbol{\tau}_t^p, \ \boldsymbol{x}_{1} = \xinit, \nonumber \\
    & && \nabla g^p \boldsymbol{\tau}_t \leq \nabla g^p \boldsymbol{\tau}_t^p - g^p,
    \tag{L-NMPC}
    \label{prob:linearized NMPC}
\end{align}
where the nonlinear cost function and constraints are Taylor-expanded at the $p$-th iteration's optimal solution $\{\boldsymbol{\tau}_t^p\}_{t= 1:T}$ to the second and first order, respectively. All the differentiation denoted by $\nabla$ is with respect to $\boldsymbol{\tau}_t$, with superscript 2 indicating the Hessian. The superscript $p$ indicates that the function is evaluated at the optimal solution of \eqref{prob:linearized NMPC} obtained in the $p$-th iteration. The solution to the displayed~\eqref{prob:linearized NMPC}, denoted by  $\{\boldsymbol{\tau}_t^{p+1}\}_{t= 1:T}$, is used in the Taylor expansion to form the $(p+1)$-th iteration of SQP.

Problem~\eqref{prob:linearized NMPC} has the identical form to that of~\eqref{prob: LQR with linear inequalities}. When the SQP terminates at the $\mathcal{N}$-th iteration, we take $\{\boldsymbol{\tau}_t^\mathcal{N}\}_{t=1:T}$ as the optimal solution to~\eqref{prob: nonlinear MPC formulation} and use~\eqref{prob: alt problem to solve for the analytical gradient} to compute the gradients. Leveraging the chain rule, we can get the target gradients $\partial \boldsymbol{u}_1^\star / \partial \boldsymbol{\theta}$ and $\partial \boldsymbol{u}_1^\star / \partial \xinit$ for DiffTune.


\begin{remark}
    For the cost function of~\eqref{prob: nonlinear MPC formulation}, the quadratic cost with $(Q_t,R_t)$-parametrization in Section~\ref{subsec: quadratic cost linear MPC} is often used for an interpretable cost function. In this case, $\nabla^2 \mathcal{C}_{\boldsymbol{\theta},t}^\mathcal{N}$ and $(\nabla \mathcal{C}_{\boldsymbol{\theta},t}^\mathcal{N} - \nabla^2 \mathcal{C}_{\boldsymbol{\theta},t}^\mathcal{N} \boldsymbol{\tau}_t^\mathcal{N})$ in the cost function of \eqref{prob:linearized NMPC} reduce to $2 \text{diag}(Q_t,R_t)$ and $-2[(
        Q_t \bar{\boldsymbol{x}}_t^\mathcal{N})^\top  \ (R_t \bar{\boldsymbol{u}}_t^\mathcal{N})^\top]^\top$, respectively. Correspondingly, the computation of $\partial \boldsymbol{u}_1^\star / \partial [Q_t]_{i,j}$ and $\partial \boldsymbol{u}_1^\star / \partial [R_t]_{i,j}$ reduces to the case covered in Section~\ref{subsec: quadratic cost linear MPC}.
\end{remark}

\section{Simulation results}\label{sec: simulation}

In this section, we implement DiffTune to learn MPCs on different dynamical systems and different trajectories. 
We compare its learning performance with a baseline learning method~\cite{song2022policy} that uses policy search (PS-MPC) to determine the parameters for MPC. We implement~\cite[Alg. 1]{song2022policy} with 10 samples of the MPC parameters in each episode. The samples follow a Gaussian distribution with the initial mean as the initial MPC parameters and the covariance matrix as $0.09I$. We use $\beta = 5$ for the inverse temperature of the soft-max distribution.

\begin{figure}
    \begin{subfigure}{0.49\columnwidth}
    \centering
    \includegraphics[width=1\columnwidth]{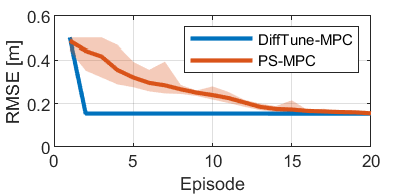} 
    \caption{Double integrator, traj.  1}   
    \label{fig: double integrator, traj 1}
\end{subfigure}
\hfill
\begin{subfigure}{0.49\columnwidth}
    \centering
    \includegraphics[width=1\columnwidth]{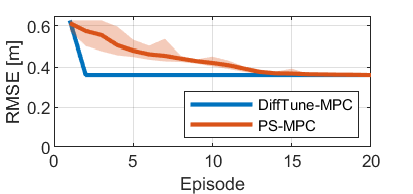} 
    \caption{Double integrator, traj.  2}   
    \label{fig: double integrator, traj 2}
\end{subfigure} \\
\begin{subfigure}{0.49\columnwidth}
    \centering
    \includegraphics[width=1\columnwidth]{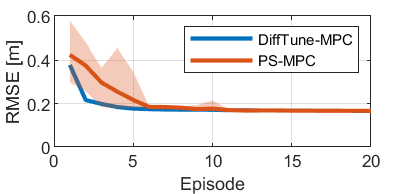} 
    \caption{Diff. wheeled robot, traj. 1}   
    \label{fig: diff wheeled robot, traj 1}
\end{subfigure}
\hfill
\begin{subfigure}{0.49\columnwidth}
    \centering
    \includegraphics[width=1\columnwidth]{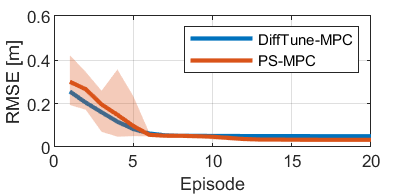} 
    \caption{Diff. wheeled robot, traj. 2}   
    \label{fig: diff wheeled robot, traj 2}
\end{subfigure}
\\
\begin{subfigure}{0.49\columnwidth}
    \centering
    \includegraphics[width=1\columnwidth]{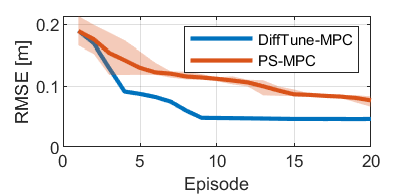} 
    \caption{Quadrotor, traj. 1}   
    \label{fig: quadrotor, traj 1}
\end{subfigure}
\hfill
\begin{subfigure}{0.49\columnwidth}
    \centering
    \includegraphics[width=1\columnwidth]{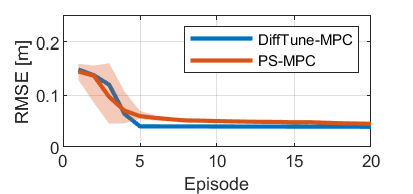} 
    \caption{Quadrotor, traj. 2}   
    \label{fig: quadrotor, traj 2}
\end{subfigure}
\caption{Simulation results comparing the learning progress of DiffTune-MPC (ours) and the baseline PS-MPC~\cite{song2022policy}. The shaded areas show the range of RMSEs (min to max) achieved with a total of 10 sampled parameters by PS-MPC.}
\label{fig: simulation}
\end{figure}

We use the following three dynamical systems: a 1D double-integrator system with box constraints for control actions and states, a differential wheeled robot with linear inequality constraints on the control actions, and a quadrotor with nonlinear dynamics. 
We use the interpretable quadratic cost function, i.e., $(Q,R)$-parametrization\footnote{We use time-invariant cost matrices in the simulation, i.e., $Q_t = Q$ and $R_t = R$ for all $t \in \{1,2,\dots,T\}$.}, within the MPC formulation in all simulation. The loss function $L$ is set to be the accumulated tracking error
squared of the closed-loop system to the reference trajectory unless specified otherwise. The $(Q,R)$ matrices are initialized as identity matrices of appropriate dimensions.
We only learn the diagonal elements of $Q$ and $R$ matrices and restrict them to be bounded within [0.01,1000].
We solve the main MPC problem using acados~\cite{Verschueren2021}. For solving~\eqref{prob: alt problem to solve for the analytical gradient}, we use acados if the original MPC problem has linear (time-invariant) dynamics and use MATLAB's \texttt{quadprog} otherwise (because acados does not support linear time-varying system of the form in \eqref{prob:linearized NMPC}). Both DiffTune and PS-MPC are set to use 20 episodes for the learning, which is sufficient for convergence with both methods (shown in Fig.~\ref{fig: simulation}).

\noindent \textbf{Double-integrator system}: Consider the following double integrator with friction
\begin{equation}
    \dot{\boldsymbol{x}}(t) = \begin{bmatrix}
        \dot{p}(t) \\ \dot{v}(t)
    \end{bmatrix} = \begin{bmatrix}
        0 & 1 \\ 0 & -0.05
    \end{bmatrix}\boldsymbol{x}(t) + \begin{bmatrix}
        0 \\ 1
    \end{bmatrix} u(t). \label{eq: double integrator with friction}
\end{equation}
We consider the boxed constraint $|u| \leq 0.9$ for control constraint and speed limit $|v| \leq 0.8$ as the state constraint. 
Two trajectories are used: $\bar{\boldsymbol{x}}_1(t) = [1+t-\cos(t), \ 1 + \sin(t)]^\top$ and $\bar{\boldsymbol{x}}_2(t) = [0.8t, \ 0.8]^\top$.
A total horizon of 10 s is considered with time intervals of 0.01~s for discretization. The planning horizon of MPC is 0.2 s ($T=20$). The learning rate is set to $\alpha = 0.5$ for DiffTune. 

\noindent \textbf{Differential wheeled robot:} Consider the following model
\begin{align}
    \dot{p}_x = \cos(\phi)u_s,\    \dot{p}_y= \sin(\phi)u_s,\    \dot{\phi} = u_\omega,
    \label{eq:unicycle}
\end{align}
where the system state $\boldsymbol{x}=[p_x,p_y,\phi]^\top$ includes the horizontal position $p_x$, vertical position $p_y$, and heading angle $\phi$, and the control action $\boldsymbol{u}=[u_s,u_\omega]^\top$ includes the speed $u_s$ and angular velocity $u_\omega$. 
The differential wheels (radius $r=0.1$ m) are placed with distance $d=0.5$~m in between. 
The maximum angular velocity for each wheel is set to $\omega_{m}=2\pi$ rad/s. Considering the geometry and the angular velocity limit of each wheel above, the following linear inequalities on the control action $\boldsymbol{u}$ apply:
\begin{align}
\label{eq:unicycle constraint}
    -2\omega_mr \le 2u_s+u_wd\le2\omega_mr, \\
    -2\omega_mr \le 2u_s-u_wd\le2\omega_mr.
\end{align}
We use the following two reference trajectories for tracking: 
\begin{align*}
    \bar{\boldsymbol{x}}_1(t) = & [1-\cos(0.5t),\ 0.5t,\ \arctan(0.5/{\sin(0.5t)})]^\top,\\
    \bar{\boldsymbol{x}}_2(t) = & [-\sin(0.5t),\ \-1+\cos(0.5t),\ 0.25t]^\top.
\end{align*}
The NMPC problem has a planning horizon of 0.5 s with 0.05 s sample time ($T=10$). 
A total horizon of $10$ s ($N=200$) is considered for the closed-loop system. We set the learning rate to $\alpha=0.01$ for DiffTune. 

\noindent \textbf{Quadrotor}: Consider the following nonlinear dynamics:
\begin{align}
    \dot{\boldsymbol{p}} = &\boldsymbol{v}, \ &&\dot{\boldsymbol{v}} =  m^{-1} f_T \boldsymbol{z}_B  + \boldsymbol{g} ,  \\ \dot{\boldsymbol{q}} =& \frac{1}{2} \boldsymbol{q} \otimes [
            0 \ \boldsymbol{\omega}^\top]^\top , \ && \dot{\boldsymbol{\omega}} = 
J^{-1} (\boldsymbol{M} - \boldsymbol{\omega} \times J \boldsymbol{\omega}),
\end{align}
where the state $\boldsymbol{x} = [        {\boldsymbol{p}}^\top\ {\boldsymbol{v}}^\top \ {\boldsymbol{q}}^\top \ {\boldsymbol{\omega}}^\top
]^\top $, $\boldsymbol{p} \in \mathbb{R}^3$, $\boldsymbol{v} \in \mathbb{R}^3$ denote the position and velocity of the quadrotor, both represented in the inertial frame, $\boldsymbol{q} \in \mathbb{S}^3$ is the quaternion representing the rotation from the inertial frame to the body frame (with $\otimes$ denoting the quaternion product), and $\boldsymbol{\omega} \in \mathbb{R}^3$ represents the angular velocity in body frame. The gravitational acceleration is denoted by $\boldsymbol{g} = [0\ 0\ -g]^\top$ (under the ENU coordinates). The vector $\boldsymbol{z}_B$ is the unit vector aligning with the $z$-axis of the body frame, represented in the inertial frame. The control $\boldsymbol{u}$ contains four individual rotor thrusts and thus determines the total thrust ${f_T}$ and moment $\boldsymbol{M} \in \mathbb{R}^3$. The quadrotor has parameters $m=0.03$~kg and $J=10^{-5}\text{diag}(1.43,1.43, 2.89)$~kgm$^2$. 
We command two reference trajectories: one is a 3D Lissajous trajectory, and the other one is a polynomial trajectory generated using minimum-snap optimization~\cite{mellinger2011minimum}. 
For both trajectories, the NMPC problem has a planning horizon of 0.5 s with 0.05 s sample time ($T=10$).
A total horizon of $10$ s ($N=200$) is considered for the closed-loop system. We set the learning rate to $\alpha=0.08$ for DiffTune. 

The results are shown in Fig.~\ref{fig: simulation}, where we display the root-mean-squared error as an indicator of the systems' performance through the learning episodes. Overall, DiffTune-MPC results in a better performance in fewer episodes than the PS-MPC, owing to the former's usage of the first-order information of the loss function. On the contrary, PS-MPC requires sampled parameters and queries to the loss function to update its probabilistic model, where a relatively large variation of RMSE happens in the first few episodes in all test cases. For the linear system, DiffTune-MPC takes one episode to reduce the RMSE too close to the minimum (the parameters are updated with small changes in the following episodes, resulting in a minor reduction in the RMSE towards the end).

\section{Experimental Results with a High-fidelity Quadrotor Simulator} \label{sec:experiment}

To show the applicability of the proposed DiffTune-MPC framework to real-world scenarios and its generalization capability (not overfitting to any particular run or condition), we conduct experiments using the high-fidelity quadrotor simulator RotorPy \cite{folk2023rotorpy}, which provides realistic aerodynamic effects, motor dynamics, and sensor noise similar to what a quadrotor experiences in physical experiments. 
Using RotorPy, we apply DiffTune for batch learning an NMPC for a trajectory tracking task subject to trajectories with different speeds. 
In particular, the task is to track 3D Lissajous reference trajectories: $\bar{\boldsymbol{p}}(t) = [ \gamma\sin(2\pi t/\mathcal{T}), \gamma\sin(4\pi t/\mathcal{T}), \gamma\sin(\pi t/\mathcal{T})]^\top \text{ for } t \in [0 , \;\mathcal{T}]$ with radius $\gamma$ and total horizon $\mathcal{T}$.
We use a total of 11 such trajectories that share the same length (fixed $\gamma$) and shape but with various speeds. We manage different speeds by setting the total horizons $\mathcal{T}$ for the 11 trajectories to be evenly distributed from 5 to 15 s. In other words, each reference trajectory is designed to finish one round within the designated horizon to achieve different desired speeds. In each training episode, we control the quadrotor with an NMPC to run these reference trajectories, compute the gradient $\nabla_{\boldsymbol{\theta}}L$ for each run, and update the parameters $(Q,R)$ using the averaged gradient from the 11 runs. The initial $(Q,R)$ for this section can be found in Appendix~G. In RotorPy, we set the simulation rate to 200 Hz and use the default vehicle and noise settings with a hummingbird profile. The NMPC problem has a planning horizon of 0.5~s with a sampling time of 0.05~s. We apply DiffTune for 15 episodes with a learning rate set to $0.01$.

\begin{figure}
\centering
\includegraphics[width=1\columnwidth]{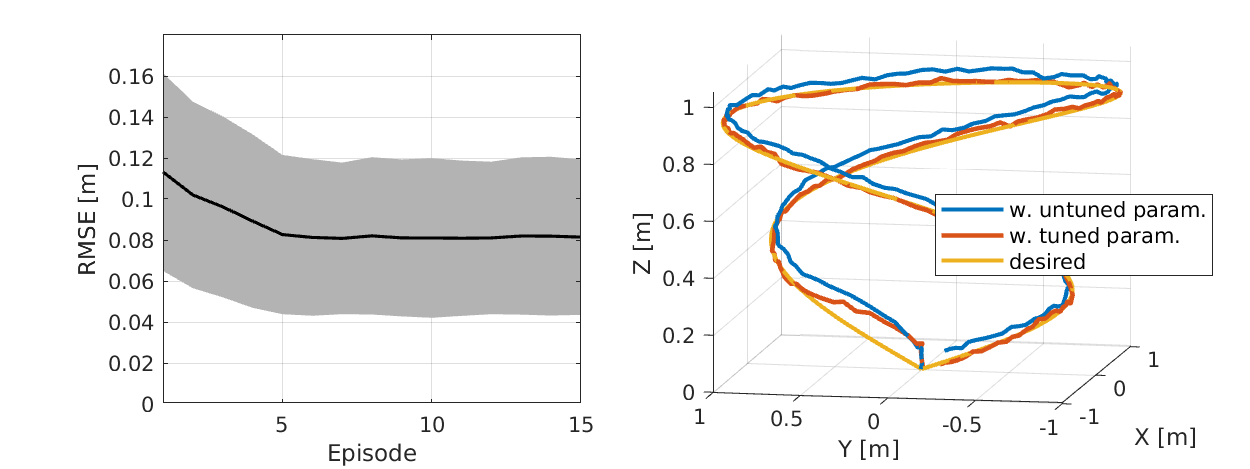} 
\caption{Left: Evolution of the tracking performance with DiffTune-MPC in RotorPy over a total of 11 3D Lissajous training trajectories. The curve shows the mean RMSE, whereas the shaded area shows the standard deviation. Right: Performance comparison over the 3D Lissajous trajectory with a total duration of 10 s.
}   
\label{fig: 2_fig}
\vspace{-5mm}
\end{figure}

The training results are shown in Fig.~\ref{fig: 2_fig}. 
It can be observed that the mean RMSE over the 11 reference trajectories decreases from 0.113 m to a final value of 0.082 m, indicating the learning capability of DiffTune in a batch-setting over multiple trajectories. 
We show the comparison of performance using learned and initial parameters in the same figure, where the improved tracking quality is clear. 
We evaluate the learned parameters over test trajectories that are not present in the training set to understand the generalization capability of the learned parameters. We also include the comparison to baseline hand-tuned parameters (these parameters can be found in Appendix~G).
The results are shown in Table \ref{table:comp_W}. The testing trajectories include 3D Lissajous trajectories with total horizons $\mathcal{T}$ outside the training set of [5,15], 3D Lissajous trajectories with different radius $\gamma$, 3D circle trajectories with different speeds, and two polynomial trajectories generated using minimum-snap optimization \cite{mellinger2011minimum}.
For the two polynomial trajectories, their average speeds are set to 1 and 2~m/s. The waypoints used to define these two trajectories can be found in Appendix~G.

Overall, the results show that the learned parameters by DiffTune-MPC outperform the hand-tuned parameters in all tested trajectories.
Over the 3D Lissajous trajectories with the same radius ($\gamma=1$), it can be observed that RMSE decreases when the total horizon increases. This tendency is expected because a larger value of the total horizon indicates a lower speed, and thus, the quadrotor can track the desired trajectory better. The same tendency can be consistently observed with 3D circle trajectories, where the RMSE decreases when the radius turns bigger (subject to a fixed total horizon), indicating a faster speed. 
Based on the training and the evaluation results shown in Fig.~\ref{fig: 2_fig} in Table \ref{table:comp_W}, we have validated the proposed framework's generalization capability for not overfitting to a particular task or condition and its applicability subject to realistic conditions with sensor noise and unknown factors (aerodynamics and rotor dynamics).



\setlength{\tabcolsep}{3pt} 
\renewcommand{\arraystretch}{1} 
  \captionsetup{
	skip=3pt, position = bottom}
\begin{table}[h]
	\centering
	\small
	
	\captionsetup{font=small}
	\caption{Evaluation of the trained parameters and the baseline hand-tuned parameters on various trajectories in RotorPy with different radius $\gamma$ and total horizon $\mathcal{T}$. The reported values are the RMSEs over 5 evaluations, displayed by mean $\pm$ standard deviation. Trajectories used in the training set are signified by *.}
    \begin{tabular}{lrr|rr}
    \toprule      
         \makecell{Trajectory}     & $\gamma$ [m]& $\mathcal{T}$ [s]          & \makecell{DiffTune-MPC  \\ RMSE [m]} & \makecell{Hand-tuned \\ RMSE [m] } \\ \midrule
3D Lissajous   &1    & 3            & \boldsymbol{$0.332 \pm 0.002$}    & $0.365 \pm 0.002$                \\ 
3D Lissajous*  &1     & 10              & \boldsymbol{$0.069 \pm 0.002$}   & $0.086 \pm 0.002$                \\ 
3D Lissajous   &1    & 20              & \boldsymbol{$0.033 \pm 0.001$}            & $0.046 \pm 0.002$               \\ 
3D Lissajous   &3    & 10    &\boldsymbol{$0.186 \pm 0.002$}            & $0.234 \pm 0.003$                \\ 
3D Lissajous  &5     & 10        & \boldsymbol{$0.301 \pm 0.001$}             & $0.395\pm 0.004$                \\ 
3D Circle &1 &10       & \boldsymbol{$0.194 \pm 0.001$}             & $0.206 \pm 0.002$                \\ 
3D Circle&3  & 10       & \boldsymbol{$0.583 \pm 0.002$}             & $0.594 \pm 0.001$               \\ 
Polynomial 1  &N/A & 10 & \boldsymbol{$0.050 \pm 0.001$}             & $0.076 \pm 0.001$\\ 
Polynomial 2  &N/A & 10 & \boldsymbol{$0.086 \pm 0.002$}             & $0.097 \pm 0.001$\\ 
        \bottomrule[1pt]
	\end{tabular}
\label{table:comp_W}
\end{table}
\normalsize

\section{Conclusion} \label{sec: conclusion}
In this paper, {we develop a closed-loop learning scheme for MPC using DiffTune.} 
We {learn} the parameters in the cost function to improve the closed-loop system's performance specified by a loss function that is different than the MPC's cost function. To use DiffTune, we need the gradient of the first optimal control action with respect to the initial state and parameters of the cost function in the MPC problem. 
Our derivation is centered on a linear MPC problem with linear inequality constraints (to incorporate safety requirements of physical limitations of the system). We show that the linear-MPC-based differentiation can be applied to a nonlinear MPC that is solved via SQP. We validate our approach in simulation for systems with linear and nonlinear dynamics subject to constraints on the state and control. Our approach shows more efficient learning compared with a baseline policy-search-based learning scheme due to our effective usage of first-order information. We also demonstrate the generalization capability of the proposed scheme to different trajectories and its applicability to real-world scenarios in a quadrotor high-fidelity simulator.
Future work will investigate the usage of DiffTune for different evaluation criteria (e.g., robustness) and the application of DiffTune-MPC on challenging robotic tasks, e.g., designing cost functions for trajectory tracking with hybrid systems in contact-rich scenarios.

\section*{Acknowledgement}
We gratefully acknowledge Yuliang Gu and Chengyu Yang for sharing their NMPC implementation in RotorPy with us.

\bibliographystyle{IEEEtran}
\bibliography{ref}

\newpage
\section*{Appendix}
\label{sec:appendix}

We show how the auxiliary problem~\eqref{prob: alt problem to solve for the analytical gradient} is derived using the KKT condition in the Appendix. We start with an LQR and then cover the case with a general linear MPC (an LQR with linear inequality constraints).
\subsection{Differentiation of an LQR}\label{subsec: LQR derivations}
We start the illustration of obtaining the gradient of an LQR's first optimal control action to the parameters of interest (cost coefficients and initial state). Consider the following LQR problem, where we stick to the same notation as in~\cite{amos2018differentiable}:
\begin{align}\label{prob: LQR}
    & \underset{\boldsymbol{\tau}_{1:T}}{\text{minimize}} && \sum_{t=1}^T \frac{1}{2}\boldsymbol{\tau}_t^\top C_t \boldsymbol{\tau}_t + \boldsymbol{c}_t^\top \boldsymbol{\tau}_t \nonumber \\
    & \text{subject to} && \boldsymbol{x}_{t+1} = F_t \boldsymbol{\tau}_t + \boldsymbol{f}_t, \ \boldsymbol{x}_1 = \xinit,\nonumber
    \tag{P1}
\end{align}
where $\boldsymbol{\tau}_t = [\begin{smallmatrix}
   \boldsymbol{x}_t \\ \boldsymbol{u}_t
\end{smallmatrix}] \in \mathbb{R}^{n+m}$ is the composite state-control, $F_t = [\begin{smallmatrix}
    A_t & B_t
\end{smallmatrix}] \in \mathbb{R}^{n \times (n+m)}$, and $\boldsymbol{f}_t \in \mathbb{R}^{n}$ stands for residuals in the dynamics.
With the Lagrangian 
\begin{align}\label{eq: Lagrangian with the basic form of LQR}
    \mathcal{L}(\boldsymbol{\tau},\boldsymbol{\lambda}) &= \sum_{t=1}^T \left(\frac{1}{2} \boldsymbol{\tau}^\top C_t \boldsymbol{\tau}_t  + \boldsymbol{c}_t^\top \boldsymbol{\tau}_t \right) \nonumber\\
    &+ \sum_{t=0}^{T-1} \boldsymbol{\lambda}_t^\top (F_t \boldsymbol{\tau}_t + \boldsymbol{f}_t - \boldsymbol{x}_{t+1}),
\end{align}
where $\boldsymbol{\lambda}_t \in \mathbb{R}^{n}$ is the vector of the Lagrangian multipliers associated with the dynamics constraints, and $F_0 = 0$ and $\boldsymbol{f}_0 = \xinit$ in the Lagrangian above. The solution to the LQR problem is obtained at the fixed point of the Lagrangian, i.e., when
\begin{equation}
    \nabla_{\boldsymbol{\tau}_t}\mathcal{L}(\boldsymbol{\tau}^\star,\boldsymbol{\lambda}^\star) = C_t \boldsymbol{\tau}_t^\star + \boldsymbol{c}_t + F_t^\top \boldsymbol{\lambda}_t^\star - \begin{bmatrix}
        \boldsymbol{\lambda}^\star_{t-1} \\ 0
    \end{bmatrix} = 0.
\end{equation}
Following this relation, we can derive the costate dynamics
\begin{align}
    \boldsymbol{\lambda}_{T-1}^\star = & C_{T,\boldsymbol{x}} \boldsymbol{x}_T^\star + \boldsymbol{c}_{T,\boldsymbol{x}},\\
    \boldsymbol{\lambda}_t^\star = & F_{t+1,\boldsymbol{x}}^\top \boldsymbol{\lambda}_{t+1} + C_{t+1,\boldsymbol{x}} \boldsymbol{x}_{t+1}^\star + \boldsymbol{c}_{t+1,\boldsymbol{x}},
\end{align}
for $t \in \{0,1,\dots,T-2\}$. 
The notation $C_{\cdot,\boldsymbol{x}}$, $F_{\cdot,\boldsymbol{x}}$, and $\boldsymbol{c}_{\cdot,\boldsymbol{x}}$ stands for the first $n$ columns of $C_{\cdot}$ and $F_{\cdot}$, and first $n$ rows of $\boldsymbol{c}_{\cdot}$, respectively.
The original dynamics, costate dynamics, and the initial state allow for writing the KKT factorization in~\eqref{eq: KKT factorization}, where $K$ is the KKT factorization matrix.
\begin{figure*}[b] 
\begin{equation}
    \underbrace{\begin{bmatrix}
        \begin{array}{cc|cc|c|cc}
             & -[\begin{smallmatrix}
                 I & 0
             \end{smallmatrix}] & &  & &  & \\
            -[\begin{smallmatrix}
                 I \\ 0
             \end{smallmatrix}] & C_1 & F_1^\top & &  & & \\
             & F_1 & & -[\begin{smallmatrix}
                 I & 0
             \end{smallmatrix}] & &  & \\
             & & -[\begin{smallmatrix}
                 I \\ 0
             \end{smallmatrix}] & C_2 &  & & \\
             & & & & \ddots &  &\\
             & & &  & & F_{T-1}^\top & \\
              &  & &  &  & & -[\begin{smallmatrix}
                 I & 0
             \end{smallmatrix}]\\
             &  & & & & -[\begin{smallmatrix}
                 I \\ 0
             \end{smallmatrix}] & C_T
        \end{array}
    \end{bmatrix}}_{:= K}
    \left[\begin{array}{c}
        \boldsymbol{\lambda}_0^\star \\ \boldsymbol{\tau}_1^\star \\ \hline  \boldsymbol{\lambda}_1^\star \\ \boldsymbol{\tau}_2^\star \\ \hline   \vdots \\ \hline  \boldsymbol{\lambda}_{T-1}^\star \\ \boldsymbol{\tau}_T^\star
    \end{array} \right] = \begin{bmatrix}
        -\xinit \\ -\boldsymbol{c}_1\\ -\boldsymbol{f}_1 \\ -\boldsymbol{c}_2 \\ \vdots \\-\boldsymbol{c}_{T-1} \\ -\boldsymbol{f}_{T-1} \\ -\boldsymbol{c}_T.
    \end{bmatrix}
    \label{eq: KKT factorization}
\end{equation}
\end{figure*}

To obtain the derivatives $\partial \boldsymbol{u}_1^\star / \partial \underline{C}_t$ and $\partial \boldsymbol{u}_1^\star / \partial \boldsymbol{c}_t$, we will take a look at the matrix differential of original dynamics, costate dynamics, and boundary conditions:
\begin{subequations}\label{eq: matrix differential of original dynamics, costate dynamics, and boundary conditions}
    \begin{align}
    \dd \boldsymbol{x}_{t+1} = & \dd F_t \boldsymbol{x}_t + F_t \dd \boldsymbol{x}_t + \dd \boldsymbol{f}_t,\\
    \dd \boldsymbol{\lambda}_t^\star = & \dd F_{t+1,\boldsymbol{x}}^\top \boldsymbol{\lambda}_{t+1} + F_{t+1,\boldsymbol{x}}^\top \dd \boldsymbol{\lambda}_{t+1} + \dd C_{t+1,\boldsymbol{x}} \boldsymbol{\tau}_{t+1}^\star \nonumber \\
    & + C_{t+1,\boldsymbol{x}} \dd \boldsymbol{\tau}_{t+1}^\star + \dd \boldsymbol{c}_{t+1,\boldsymbol{x}},\\
    \dd \boldsymbol{x}_1 = & \dd \xinit,\\
    \dd \boldsymbol{\lambda}_{T-1}^\star = & \dd C_{T,\boldsymbol{x}} \boldsymbol{x}_T^\star + C_{T,\boldsymbol{x}} \dd \boldsymbol{x}_T^\star + \dd \boldsymbol{c}_{T,\boldsymbol{x}}.
\end{align}
\end{subequations}

The KKT factorization for~\eqref{eq: matrix differential of original dynamics, costate dynamics, and boundary conditions} can be presented as follows:
\begin{equation} 
    K\begin{bmatrix}
        \dd \boldsymbol{\lambda}_0^\star \\ \dd \boldsymbol{\tau}_1^\star \\ \dd \boldsymbol{\lambda}_1^\star \\ \dd \boldsymbol{\tau}_2^\star \\ \vdots \\ \dd \boldsymbol{\lambda}_{T-1}^\star \\ \dd \boldsymbol{\tau}_T^\star
    \end{bmatrix} = \begin{bmatrix}
        -\dd \xinit \\ -\dd \boldsymbol{c}_1 -\dd C_1 \boldsymbol{\tau}_1^\star - \dd F_1^\top \boldsymbol{\lambda}_1^\star\\ -\dd \boldsymbol{f}_1 - \dd F_1 \boldsymbol{\tau}_1^\star \\ -\dd \boldsymbol{c}_2 - \dd C_2 \boldsymbol{\tau}_2^\star - \dd F_2^\top \boldsymbol{\lambda}_2^\star \\ \vdots \\-\dd \boldsymbol{c}_{T-1} - \dd C_{T-1} \boldsymbol{\tau}_{T-1}^\star - \dd F_{T-1}^\top \boldsymbol{\lambda}_{T-1}^\star \\ -\dd \boldsymbol{f}_{T-1} - \dd F_{T-1} \boldsymbol{\tau}_{T-1}^\star \\ -\dd \boldsymbol{c}_T - \dd C_T \boldsymbol{\tau}_T^\star
    \end{bmatrix}.
    \label{eq: KKT factorization for differentials}
\end{equation}

We start with the Jacobian $\partial \boldsymbol{u}_1^\star / \partial \xinit$ because it is simpler than the other two Jacobians. We set the differentials $\dd C_t$, $\dd \boldsymbol{c}_t$, $\dd F_t$, and $\dd \boldsymbol{f}_t$ to zero for $t \in \{1,2,\dots,T\}$ in \eqref{eq: KKT factorization for differentials} and we have
\begin{equation}
    K \begin{bmatrix}
        \frac{\partial \boldsymbol{\lambda}_0^\star}{\partial \xinit} \\
        \frac{\partial \boldsymbol{\tau}_1^\star}{\partial \xinit} \\
        \vdots \\
        \frac{\partial \boldsymbol{\lambda}_{T-1}^\star}{\partial \xinit} \\
        \frac{\partial \boldsymbol{\tau}_T^\star}{\partial \xinit} 
    \end{bmatrix} = \begin{bmatrix}
            -I_n \\ 0 \\ \vdots \\ 0 \\ 0
        \end{bmatrix}, \label{eq: example to illustrate du/dxinit using KKT factorization}
\end{equation}
where the target Jacobian $\partial \boldsymbol{u}_1^\star / \partial \xinit$ is contained within $\partial \boldsymbol{\tau}_1^\star / \partial \xinit$ such that
\begin{equation}
    \frac{\partial \boldsymbol{\tau}_1^\star}{\partial \xinit} = \begin{bmatrix}
        \frac{\partial \boldsymbol{x}_1^\star}{\partial \xinit} \\
        \frac{\partial \boldsymbol{u}_1^\star}{\partial \xinit}
    \end{bmatrix}.
\end{equation}
If we compare the form and match the terms of \eqref{eq: example to illustrate du/dxinit using KKT factorization} to the KKT factorization of the original problem in \eqref{eq: KKT factorization}, then we can obtain $\partial \boldsymbol{u}_1^\star / \partial \xinit$ by solving the following auxiliary LQR problem 
\begin{align}
\label{eq:LQR-Aux}
    & \underset{\tilde{\boldsymbol{\tau}}_{1:T}}{\text{minimize}} && \sum_{t=1}^T \frac{1}{2}\tilde{\boldsymbol{\tau}}_t^\top \tilde{C}_t \tilde{\boldsymbol{\tau}}_t + \tilde{\boldsymbol{c}}_t^\top \tilde{\boldsymbol{\tau}}_t \nonumber \\
    & \text{subject to} && \tilde{\boldsymbol{x}}_{t+1} = \tilde{F}_t \tilde{\boldsymbol{\tau}}_t + \tilde{\boldsymbol{f}}_t, \ \tilde{\boldsymbol{x}}_1 = \tilde{\boldsymbol{x}}_\text{init}.\nonumber
    \tag{LQR-Aux}
\end{align}
The gradient $\partial \boldsymbol{u}_1^\star / \partial [\xinit]_j$ can be solved from \eqref{eq:LQR-Aux} following the observations below.
By \eqref{eq: example to illustrate du/dxinit using KKT factorization}, we have
\begin{equation}
    K \begin{bmatrix}
        \frac{\partial \boldsymbol{\lambda}_0^\star}{\partial [\xinit]_j} \\
        \frac{\partial \boldsymbol{\tau}_1^\star}{\partial [\xinit]_j} \\
        \vdots \\
        \frac{\partial \boldsymbol{\lambda}_{T-1}^\star}{\partial [\xinit]_j} \\
        \frac{\partial \boldsymbol{\tau}_T^\star}{\partial [\xinit]_j} 
    \end{bmatrix} = \begin{bmatrix}
            -I_n \\ 0 \\ \vdots \\ 0 \\ 0
        \end{bmatrix}_{:,j}. \label{eq: elementwise du/dxinit using KKT factorization}
\end{equation}
Matching \eqref{eq: elementwise du/dxinit using KKT factorization} with \eqref{eq: KKT factorization}, $\partial \boldsymbol{u}_1^\star / \partial [\xinit]_j$ can be solved via \eqref{eq:LQR-Aux} with $\tilde{C}_t = C_t$, $\tilde{\boldsymbol{c}}_t = 0$, $\tilde{F}_t = F_t$ , $\Tilde{\boldsymbol{f}}_t = 0$, and $\Tilde{\boldsymbol{x}}_{\text{init}} = [I_n]_{:,j} = \boldsymbol{e}_j^{n}$.

Similarly, for the Jacobian $\partial \boldsymbol{u}_1^\star / \partial \underline{C}_t$, we start with
\begin{equation}\label{eq: matrix differential for du/dCt}
    \dd C_t \boldsymbol{\tau}_t^\star = ((\boldsymbol{\tau}_t^\star)^\top \otimes I_{n+m}) \dd \underline{C}_t
\end{equation}
using the vectorization (with Kronecker product) formula $\underline{ABC} = (C^\top \otimes A) \underline{B}$.
This equation will lead to the factorization
\begin{equation}\label{eq: KKT factorization for du/dC_t}
    K \begin{bmatrix}
        * \\
        \partial \boldsymbol{\tau}_1^\star / \partial \underline{C}_t \\
        *
    \end{bmatrix} = \begin{bmatrix}
        0 \\
        \vdots \\
        \begin{bmatrix}
            -(\boldsymbol{\tau}_t^\star)^\top \otimes I_{n+m}
        \end{bmatrix} \\
        0 \\
        \vdots
    \end{bmatrix}.
\end{equation}
Matching \eqref{eq: KKT factorization for du/dC_t} with \eqref{eq: KKT factorization}, given any specific $t$, the derivative $\partial \boldsymbol{u}_1^\star / \partial [C_t]_{i,j}$ can be solved from \eqref{eq:LQR-Aux} with $\tilde{C}_t = C_t$, $\tilde{F}_t = F_t$, $\Tilde{\boldsymbol{f}}_t = 0$ for all $t$, $\Tilde{\boldsymbol{x}}_{\text{init}} = 0$, and $\Tilde{\boldsymbol{c}}_t =[
            (\boldsymbol{\tau}_t^\star)^\top \otimes I_{n+m}
        ]_{:,(i-1)(n+m)+j}
         = [\boldsymbol{\tau}_t^\star]_j \boldsymbol{e}_i^{n+m}$ for this specific $t$ and $\Tilde{\boldsymbol{c}}_{t'}=0$ for all $t' \neq t$.
Likewise, if one wants to get $\partial \boldsymbol{u}_1^\star / \partial [\boldsymbol{c}_t]_{i}$, then it suffices to solve \eqref{eq:LQR-Aux} with $\tilde{\boldsymbol{c}}_t =  \boldsymbol{e}^{n+m}_i$ and $\Tilde{\boldsymbol{c}}_{t'}=0$ for all $ t' \neq t$.

\subsection{Differentiation of an LQR with $(Q,R)$-parametrization}
In this case, we use the $(Q,R)$-parameterization with coefficients of the $Q_t \in \mathbb{R}^{n \times n}$ and $R_t \in \mathbb{R}^{m \times m}$ matrices in the cost function, which are composited as $C_t = \text{diag}(Q_t,R_t)$ following the notation in~\eqref{prob: LQR}. Note that $\boldsymbol{c}_t$ also contains the coefficients of $Q_t$ and $R_t$ such that $\boldsymbol{c}_t = -[\begin{smallmatrix} Q_t & \\ & R_t\end{smallmatrix}] \bar{\boldsymbol{\tau}_t}$ for $\bar{\boldsymbol{\tau}}_t = [\begin{smallmatrix}
    \bar{\boldsymbol{x}}_t \\ \bar{\boldsymbol{u}}_t
\end{smallmatrix}]$. 

For $\partial \boldsymbol{u}_1^\star / \partial \underline{Q}_t$, one can apply the same methodology as applied in the previous subsection. Note that
\begin{align}
    \dd C_t \boldsymbol{\tau}_t^\star + \dd \boldsymbol{c}_t = &\begin{bmatrix}
         \dd Q_t(\boldsymbol{x}_t^\star - \bar{\boldsymbol{x}}_t)\\\dd R_t(\boldsymbol{u}_t^\star - \bar{\boldsymbol{u}}_t) 
     \end{bmatrix}
      \\
    = & \begin{bmatrix}
\left((\boldsymbol{x}_t^\star - \bar{\boldsymbol{x}}_t)^\top \otimes I_n \right) \dd \underline{Q}_t \\
\left((\boldsymbol{u}_t^\star - \bar{\boldsymbol{u}}_t)^\top \otimes I_{m} \right) \dd \underline{R}_t
    \end{bmatrix}
\end{align}
If $Q_t$ is symmetric, then $\dd Q_t(\boldsymbol{x}_t^\star - \bar{\boldsymbol{x}}_t) = \dfrac{1}{2}\left((\boldsymbol{x}_t^\star - \bar{\boldsymbol{x}}_t)^\top \otimes I_{n} + I_{n} \otimes (\boldsymbol{x}_t^\star - \bar{\boldsymbol{x}}_t)^\top \right) \dd \underline{Q}_t$ (the same conclusion applies to a symmetric matrix $R_t$).  Hence, we have
\begin{equation}
    K \begin{bmatrix}
        * \\
        \partial \boldsymbol{\tau}_1^\star / \partial [{Q}_t]_{i,j} \\
        *
    \end{bmatrix} = \begin{bmatrix}
        0 \\
        \vdots \\
        \begin{bmatrix}
            -(\boldsymbol{x}_t^\star - \bar{\boldsymbol{x}}_t)^\top \otimes I_n\\0
        \end{bmatrix} \\
        0 \\
        \vdots
    \end{bmatrix}_{:,(i-1)n+j}.
\end{equation}
In other words, the derivative $\partial \boldsymbol{u}_1^\star / \partial [Q_t]_{i,j}$ can be solved from \eqref{eq:LQR-Aux} with $\tilde{C}_t = C_t$, $\Tilde{F}_t = F_t$, $\Tilde{\boldsymbol{f}}_t = 0$, $\Tilde{\boldsymbol{x}}_{\text{init}} = 0$ for all $t$, and 
$\Tilde{\boldsymbol{c}}_t =
\begin{bmatrix}
            (\boldsymbol{x}_t^\star - \bar{\boldsymbol{x}}_t)^\top \otimes I_n\\0
        \end{bmatrix}_{:,(i-1)n+j} = 
        \begin{bmatrix}
            [\boldsymbol{x}_t^\star - \bar{\boldsymbol{x}}_t]_j \boldsymbol{e}_i^{n}\\0
        \end{bmatrix}
        $
for this specific $t$ and $\tilde{\boldsymbol{c}}_{t'}=0$ for all $t' \neq t$. One can apply similar derivations to obtain $\partial \boldsymbol{u}_1^\star / \partial [R_t]_{i,j}$ from \eqref{eq:LQR-Aux} with $\Tilde{\boldsymbol{x}}_{\text{init}} = 0$ for all $t$, and
$\Tilde{\boldsymbol{c}}_t =
\begin{bmatrix}
            0\\(\boldsymbol{u}_t^\star - \bar{\boldsymbol{u}}_t)^\top \otimes I_n
        \end{bmatrix}_{:,(i-1)n+j} = 
        \begin{bmatrix}
            0\\ [\boldsymbol{u}_t^\star - \bar{\boldsymbol{u}}_t]_j \boldsymbol{e}_i^m
        \end{bmatrix}$ and $\tilde{\boldsymbol{c}}_{t'}=0$ for all $t' \neq t$.

\subsection{Differentiation of an LQR with linear inequality constraints}
\label{sec:LQR with LI}
A slightly enhanced version of LQR includes path constraints on the state and control, e.g., with linear inequalities, which results in the formulation of~\eqref{prob: LQR with linear inequalities}. Following the derivation in Appendix A, we can write the Lagrangian for~\eqref{prob: LQR with linear inequalities} as
\begin{align}\label{eq: Lagrangian with LQR-LI}
     & \mathcal{L}(\boldsymbol{\tau},\boldsymbol{\lambda},\boldsymbol{\nu}) \nonumber \\
    =& \sum_{t=1}^T \left(\frac{1}{2} \boldsymbol{\tau}_t^\top C_t \boldsymbol{\tau}_t  + \boldsymbol{c}_t^\top \boldsymbol{\tau}_t \right) + \sum_{t=0}^{T-1} \boldsymbol{\lambda}_t^\top (F_t \boldsymbol{\tau}_t + \boldsymbol{f}_t - \boldsymbol{x}_{t+1}) \nonumber \\
    & + \sum_{t=1}^T \boldsymbol{\nu}_t^\top (G_t\boldsymbol{\tau}_t - \boldsymbol{l}_t),
\end{align}
where $\boldsymbol{\nu}_t \in \mathbb{R}^g$ is the Lagrange multiplier associated with the inequality constraints. 
Let $\mathcal{I}_{t,\text{act}} \subseteq \{1,2,\dots, g\}$ denote the rows where the inequality $G_t \boldsymbol{\tau}_t \leq \boldsymbol{l}_t$ is active, i.e., $[G_t]_{\mathcal{I}_{t,\text{act}}} \boldsymbol{\tau}_t = [\boldsymbol{l}_t]_{\mathcal{I}_{t,\text{act}}}$. Correspondingly, $\mathcal{I}_{t,\text{ina}} = \{1,2,\dots, g\} \setminus \mathcal{I}_{t,\text{act}}$ denotes the inactive constraints, i.e., $[G_t]_{\mathcal{I}_{t,\text{ina}}} \boldsymbol{\tau}_t < [\boldsymbol{l}_t]_{\mathcal{I}_{t,\text{ina}}}$
Following the complementary slackness~\cite{boyd2004convex}, we know the following condition holds at the optimality point:
\begin{align}
    & [\boldsymbol{\nu}_t]_{\mathcal{I}_{t,\text{ina}}} = 0, \quad [G_t]_{\mathcal{I}_{t,\text{ina}}} \boldsymbol{\tau}_t - [\boldsymbol{l}_t]_{\mathcal{I}_{t,\text{ina}}}< 0, \\
    & [G_t]_{\mathcal{I}_{t,\text{act}}} \boldsymbol{\tau}_t - [\boldsymbol{l}_t]_{\mathcal{I}_{t,\text{act}}}= 0, \quad [\boldsymbol{\nu}_t]_{\mathcal{I}_{t,\text{act}}} > 0.
\end{align}

The costate dynamics for \eqref{prob: LQR with linear inequalities} are
\begin{align}
    \boldsymbol{\lambda}_{T-1}^\star = & C_{T,\boldsymbol{x}} \boldsymbol{x}_T^\star + \boldsymbol{c}_{T,\boldsymbol{x}} + G_{T,\boldsymbol{x}}^\top \boldsymbol{\nu}_T^\star,\\
    \boldsymbol{\lambda}_t^\star = & F_{t+1,\boldsymbol{x}}^\top \boldsymbol{\lambda}_{t+1} + C_{t+1,\boldsymbol{x}} \boldsymbol{x}_{t+1}^\star + \boldsymbol{c}_{t+1,\boldsymbol{x}}+ G_{t,\boldsymbol{x}}^\top \boldsymbol{\nu}_t^\star.
\end{align}
We can write down the KKT factorization for~\eqref{prob: LQR with linear inequalities} as shown in~\eqref{eq: KKT factorization for LQR-LI}, with the KKT factorization matrix denoted by $K_{LI}$.

\begin{figure*}
\begin{equation}
    \underbrace{\begin{bmatrix}
        \begin{array}{ccc|ccc|c|ccc}
             & -[\begin{smallmatrix}
                 I & 0
             \end{smallmatrix}] & & & & &  & &  &  \\
            -[\begin{smallmatrix}
                 I \\ 0
             \end{smallmatrix}] & C_1 & G_1^\top &  F_1^\top & & & & &  &  \\
             & & I_{\mathcal{I}_{1,\text{ina}}} & & & & & & &  \\
             & [G_1]_{\mathcal{I}_{1,\text{act}}} &  & & & & & & &  \\
             & F_1 & & & -[\begin{smallmatrix}
                 I & 0
             \end{smallmatrix}] & &  & & &  \\
             & & & -[\begin{smallmatrix}
                 I \\ 0
             \end{smallmatrix}] & C_2 & G_2^\top &  & &  \\
             & & & & & I_{\mathcal{I}_{2,\text{ina}}} & & & &  \\
             & & & & [G_2]_{\mathcal{I}_{2,\text{act}}} & & & & &  \\
             & & & & & & \ddots & & &  \\
             & & &  & &  &  & & -[\begin{smallmatrix}
                 I & 0
             \end{smallmatrix}] &  \\
             & & &  & & & & -[\begin{smallmatrix}
                 I \\ 0
             \end{smallmatrix}] & C_T & G_T^\top \\
             & & &  & & & &  &  &  I_{\mathcal{I}_{T,\text{ina}}} \\
             & & &  & & & &  & [G_T]_{\mathcal{I}_{T,\text{act}}} &   \\
        \end{array}
    \end{bmatrix}}_{:= K_{LI}}
    \left[ \begin{array}{c}
        \boldsymbol{\lambda}_0^\star \\ \boldsymbol{\tau}_1^\star \\ \boldsymbol{\nu}_1^\star \\ \hline \boldsymbol{\lambda}_1^\star \\ \boldsymbol{\tau}_2^\star \\ \boldsymbol{\nu}_2^\star \\ \hline  \vdots \\ \hline  \boldsymbol{\lambda}_{T-1}^\star \\ \boldsymbol{\tau}_T^\star \\ \boldsymbol{\nu}_t^\star
    \end{array} \right] = \begin{bmatrix}
        -\xinit \\ -\boldsymbol{c}_1\\ 0_{\mathcal{I}_{1,\text{ina}}} \\ [\boldsymbol{l}_1]_{\mathcal{I}_{1,\text{act}}} \\ -\boldsymbol{f}_1 \\ -\boldsymbol{c}_2 \\ 0_{\mathcal{I}_{2,\text{ina}}} \\ [\boldsymbol{l}_2]_{\mathcal{I}_{1,\text{act}}} \\ \vdots \\ -\boldsymbol{f}_{T-1} \\ -\boldsymbol{c}_T \\ 0_{\mathcal{I}_{T,\text{ina}}} \\ [\boldsymbol{l}_t]_{\mathcal{I}_{T,\text{act}}}.
    \end{bmatrix}
    \label{eq: KKT factorization for LQR-LI}
\end{equation}
\end{figure*}

For the differentials associated with~\eqref{prob: LQR with linear inequalities}, we have
\begin{equation}
    K_{LI}\begin{bmatrix}
        \dd \boldsymbol{\lambda}_0^\star \\ \dd \boldsymbol{\tau}_1^\star \\ \dd \boldsymbol{\nu}_1^\star \\  \vdots \\  \dd \boldsymbol{\lambda}_{T-1}^\star \\ \dd \boldsymbol{\tau}_T^\star \\ \dd \boldsymbol{\nu}_T^\star
    \end{bmatrix} = \begin{bmatrix}
        -\dd \xinit \\ -\dd \boldsymbol{c}_1 -\dd C_1 \boldsymbol{\tau}_1^\star - \dd F_1^\top \boldsymbol{\lambda}_1^\star -\dd G_1^\top \boldsymbol{\nu}_1^\star \\ 0_{\mathcal{I}_{1,\text{ina}}} \\ -\dd [G_1]_{\mathcal{I}_{1,\text{act}}} \boldsymbol{\tau}_1^\star + \dd [\boldsymbol{l}_1]_{\mathcal{I}_{1,\text{act}}} \\ \vdots \\ -\dd \boldsymbol{f}_{T-1} - \dd F_{T-1} \boldsymbol{\tau}_{T-1}^\star \\ -\dd \boldsymbol{c}_T - \dd C_T \boldsymbol{\tau}_T^\star - \dd G_T^\top \boldsymbol{\nu}_T^\star \\ 0_{\mathcal{I}_{1,\text{ina}}} \\ -\dd [G_T]_{\mathcal{I}_{T,\text{act}}} \boldsymbol{\tau}_1^\star + \dd [\boldsymbol{l}_t]_{\mathcal{I}_{T,\text{act}}}
    \end{bmatrix}
    \label{eq: KKT factorization for differentials of LQR-LI}
\end{equation}
One observation follows from~\eqref{eq: KKT factorization for differentials of LQR-LI}: when considering the Jacobians with respect to $C_t$ and $\boldsymbol{c}_t$, then only non-zero differentials on the right-hand side of~\eqref{eq: KKT factorization for differentials of LQR-LI} will be $\dd C_t$ and $\dd \boldsymbol{c}_t$, which resembles the structure of the right-hand side of~\eqref{eq: KKT factorization for differentials}. This observation indicates that the gradients $\partial \boldsymbol{u}_1^\star / \partial \underline{C}_t$ and $\partial \boldsymbol{u}_1^\star / \partial \boldsymbol{c}_t$ can be obtained following the approach shown before with LQR in Appendix~A. 

Following the derivation in Appdendix~A for \eqref{prob: LQR}, one can solve the following auxiliary problem to obtain the gradient for \eqref{prob: LQR with linear inequalities}:
\begin{align}
    & \underset{\tilde{\boldsymbol{\tau}}_{1:T}}{\text{minimize}} && \sum_{t=1}^T \frac{1}{2}\tilde{\boldsymbol{\tau}}_t^\top \tilde{C}_t \tilde{\boldsymbol{\tau}}_t + \tilde{\boldsymbol{c}}_t^\top \tilde{\boldsymbol{\tau}}_t \nonumber \\
    & \text{subject to} && \tilde{\boldsymbol{x}}_{t+1} = \tilde{F}_t \tilde{\boldsymbol{\tau}}_t+\tilde{\boldsymbol{f}}_t , \ \tilde{\boldsymbol{x}}_1 = \tilde{\boldsymbol{x}}_{\text{init}}, \nonumber \\
    & && \tilde{G}_t \tilde{\boldsymbol{\tau}}_t = \tilde{\boldsymbol{l}}_t.
    \nonumber
    \tag{LMPC-Aux}
    \label{prob: LMPC-aux}
\end{align}
Specifically, to obtain the gradient $\partial \boldsymbol{u}_1^\star / \partial [C_t]_{i,j}$, one only needs to solve \eqref{prob: LMPC-aux} with $\tilde{C}_t = C_t$, $\Tilde{\boldsymbol{c}}_t =
\begin{bmatrix}
    (\boldsymbol{\tau}_1^\star)^\top \otimes I_{n+m}
\end{bmatrix}_{:,(i-1)(n+m)+j} = [\boldsymbol{\tau}_t^\star]_j \boldsymbol{e}_i^{n+m}$, $\Tilde{\boldsymbol{f}}_t = 0$, $\Tilde{\boldsymbol{x}}_{\text{init}} = 0$, $\Tilde{G}_t = [G_t]_{\mathcal{I}_{t,\text{act}}}$, and $\Tilde{\boldsymbol{l}}_t = 0$. Likewise, to solve for the gradient $\partial \boldsymbol{u}_1^\star / \partial [\boldsymbol{c}_t]_i$, we can solve the problem \eqref{prob: LMPC-aux} with $\tilde{C}_t = C_t$, $\Tilde{\boldsymbol{c}}_t = \boldsymbol{e}_i^{n+m}$, $\Tilde{\boldsymbol{f}}_t = 0$, $\Tilde{\boldsymbol{x}}_{\text{init}} = 0$, $\Tilde{G}_t = [G_t]_{\mathcal{I}_{t,\text{act}}}$, and  $\Tilde{\boldsymbol{l}}_t = 0$. 
To solve for $\partial \boldsymbol{u}_1^\star / \partial [\xinit]_j$, one can solve \eqref{prob: LMPC-aux} with $\tilde{C}_t = C_t$, $\Tilde{\boldsymbol{c}}_t = 0$, $\Tilde{\boldsymbol{f}}_t = 0$, $\Tilde{\boldsymbol{x}}_{\text{init}} = \boldsymbol{e}_j^n$, $\Tilde{G}_t = [G_t]_{\mathcal{I}_{t,\text{act}}}$, and $\Tilde{\boldsymbol{l}}_t = 0$. These results are summarized in \eqref{prob: alt problem to solve for the analytical gradient} and Table~\ref{tb: table for gradient computation with C c parametrization} in the text.

Note that the inequality constraints $G_t \boldsymbol{\tau}_t \leq \boldsymbol{l}_t$ in \eqref{prob: LQR with linear inequalities} are turned into an equality constraint $\tilde{G}_t \tilde{\boldsymbol{\tau}}_t = \tilde{\boldsymbol{l}}_t$ in \eqref{prob: LMPC-aux}, where the latter equality constraint only holds for those time indices $t$ with the constraint being active at the optimal solution $\boldsymbol{\tau}_t^\star$, i.e., $\tilde{G}_t = [G_t]_{\mathcal{I}_{t,\text{act}}}$. The inactive constraints (at the optimal solution of \eqref{prob: LQR with linear inequalities}) are not considered in \eqref{prob: LMPC-aux} because the optimal solution of \eqref{prob: LQR with linear inequalities} is ``locally unconstrained'' at the inactive constraints, i.e., removing the inactive constraints will not change the optimal solution. Therefore, the inactive constraints $[G_t]_{\mathcal{I}_{t,\text{ina}}}$ are not inherited from \eqref{prob: LQR with linear inequalities} to \eqref{prob: alt problem to solve for the analytical gradient}, and only active constraints show up in \eqref{prob: alt problem to solve for the analytical gradient} in the form of $\tilde{G}_t \tilde{\boldsymbol{\tau}}_t = 0$ for $ \tilde{G}_t = [G_t]_{\mathcal{I}_{t,\text{act}}}$.

\subsection{Differential wheeled robot simulation}
\begin{figure}
    \begin{subfigure}{\columnwidth}
    \centering
    \includegraphics[width=1\columnwidth]{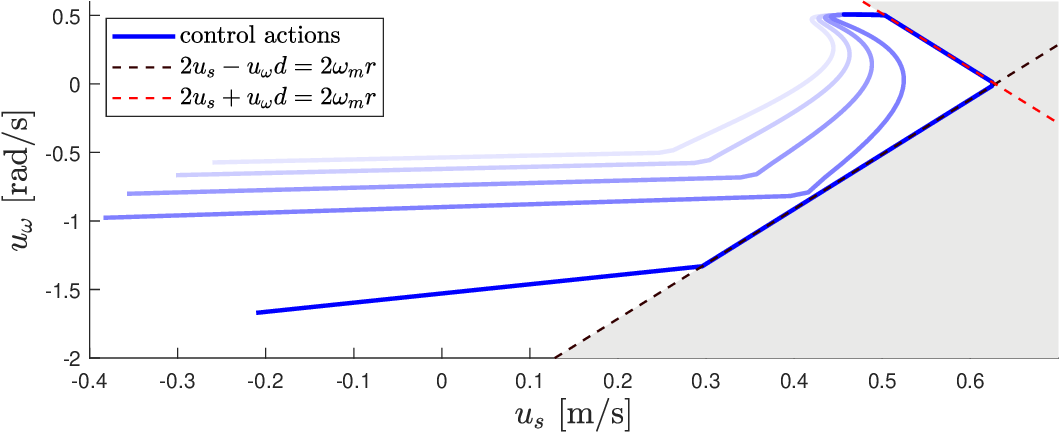} 
    \caption{Optimal closed-loop control actions through the learning episodes. The lightest and darkest blue curves stand for episodes 1 and 20, respectively.}   
    \label{fig: evolution of control, traj 2}
\end{subfigure}
\\
\begin{subfigure}{\columnwidth}
    \centering
    \includegraphics[width=1\columnwidth]{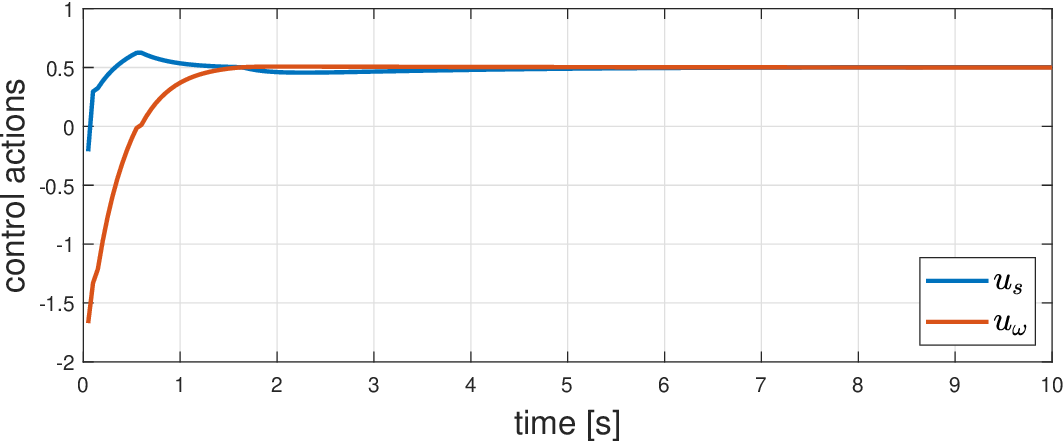} 
    \caption{Time history of the control signal in the last episode.}   
    \label{fig: control history trajec 2}
\end{subfigure}
\caption{Optimal closed-loop control actions with the differential wheeled robot under trajectory 2.}
\label{fig: opt control traj 2 for diff wheeled robot}
\end{figure}

We show the evolution of control actions through the learning episodes in Fig.~\ref{fig: opt control traj 2 for diff wheeled robot}, representing the learning conducted over trajectory 2 for the differential wheeled robot.
Towards the end of the episodes, the robot learns to use control actions on the boundary of the feasible control set (see Fig.~\ref{fig: evolution of control, traj 2}) to catch up with the desired circular trajectory and then maintain a steady state with constant forward velocity $u_s$ and rotational velocity $u_{\omega}$. The time history is shown in Fig.~\ref{fig: control history trajec 2}.

\subsection{Impact of control saturation on learning}
\label{subsec: double integrator_appendix}
\begin{figure}[h]

    \begin{subfigure}{\columnwidth}
        \includegraphics[width = \columnwidth]{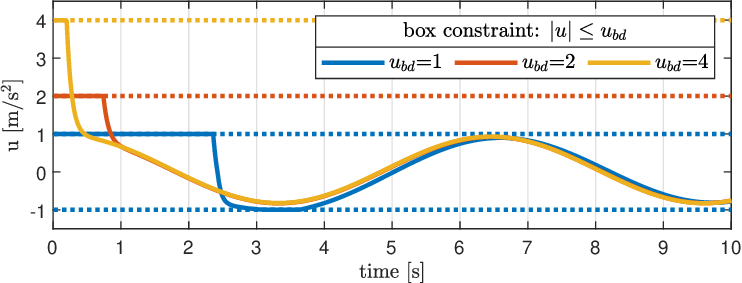}
        \caption{Closed-loop optimal control in the final episode. Dashed lines indicate the boundary $u_{bd}$ of control actions.}
        \label{fig: opt control 1D example with sat}
    \end{subfigure}
    \begin{subfigure}{\columnwidth}
        \includegraphics[width = \columnwidth]{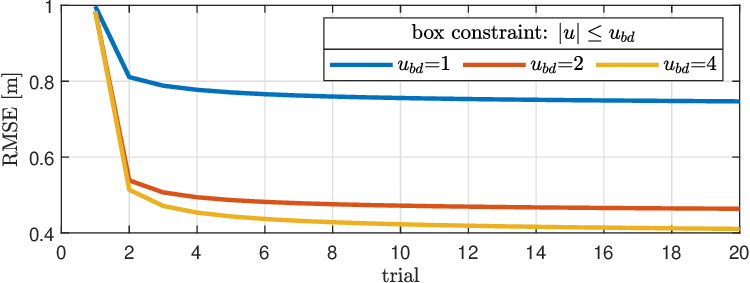}
        \caption{RMSE reduction}
        \label{fig: RMSE for 1D example with sat}
    \end{subfigure}
    \caption{Optimal control actions and RMSE reduction subject to box constraints with different levels of tightness $u_{bd}$. Tighter constraints imply more frequent control saturation, which impedes learning from improving tracking performance.}
  \label{fig:1D example to show the impact of sat to tuning}

  \vspace{-0.6 cm}
\end{figure}

We use a simple linear system to illustrate the impact of control saturation on learning. Consider a 1D double-integrator system used in Section~\ref{sec: simulation}.
We command the following reference trajectory for tracking $\bar{\boldsymbol{x}}(t) = [1+t-\cos(t), \ 1 + \sin(t)]^\top$. We consider the boxed constraint $|u| \leq u_{bd}$ for control constraints and remove the state constraints. We choose $u_{bd}=1,2,4$ to reflect the different magnitudes of constrained control actions. A total horizon of 10 s is considered with time intervals of 0.01~s for discretization. The planning horizon of MPC is 0.2 s ($T=20$). We apply DiffTune to these three cases for 20 episodes each with the loss function set to the accumulated tracking error squared and learning rate $\alpha=0.1$. The optimal closed-loop control actions in the last episode are shown in Fig.~\ref{fig: opt control 1D example with sat}, where the saturation at the beginning lasts for a shorter period for a bigger value of $u_{bd}$. Correspondingly, the tracking error has a larger reduction for larger values of $u_{bd}$ as shown in Fig.~\ref{fig: RMSE for 1D example with sat}. This observation validates our earlier analysis in Section~\ref{subsec: diff linear MPC}, where active constraints (saturated control) cap performance improvement by learning. Tighter constraints imply more frequent control saturation, which impedes the auto-tuning to improve tracking performance because saturation is a physical limitation that is beyond what learning is for.

\subsection{Insights from (LMPC-Grad) for general controllers subject to control constraints}\label{subsec: active constraints}
    Handling the analytical gradient of \eqref{prob: LQR with linear inequalities} with constraint $G_t \boldsymbol{\tau}_t \leq \boldsymbol{l}_t$ using $[G_t]_{\mathcal{I}_{t,\text{act}}} \tilde{\boldsymbol{\tau}}_t = 0$ in \eqref{prob: alt problem to solve for the analytical gradient} is instructional to the differentiation of control policies that have control constraints (e.g., saturation) in the form of $G \boldsymbol{u} \leq l$. Many controllers are designed without constraints, like PID. However, when these controllers are implemented on real-world physical systems, strict adherence to state and/or input constraints is often required due to the actuator limitations or safety considerations. Mathematically, these constraint-adhering controller can be formulated as $h = \mathfrak{c}_{G,l} \circ h_0$, which is a composition of the constraining operator $\mathfrak{c}_{G,l}$ and an arbitrary constraint-free policy $h_0$, where
    \begin{align}
        \mathfrak{c}_{G,l}(\boldsymbol{u}) = & \text{\ } \underset{\hat{\boldsymbol{u}}}{\text{argmin}} \text{\ } \norm{\boldsymbol{u}-\hat{\boldsymbol{u}}} \nonumber \\
        & \text{subject to} \text{\ } G \hat{\boldsymbol{u}} \leq l. \label{prob: constraints as optimization}
    \end{align}
    Following the notation of control policy in \eqref{eq: feedback controller}, we have  $h(\boldsymbol{x},\bar{\boldsymbol{x}},\boldsymbol{\theta}) = \mathfrak{c}_{G,l}(h_0(\boldsymbol{x},\bar{\boldsymbol{x}},\boldsymbol{\theta}))$ for $\bar{\boldsymbol{x}}$ being the reference state and $\boldsymbol{\theta}$ being the parameters to be learned. To obtain $\partial h / \partial \boldsymbol{\theta}$ for learning, one can apply chainrule by $\frac{\partial h}{\partial \boldsymbol{\theta}} = \frac{\partial \mathfrak{c}_{G,l}(\boldsymbol{u})}{\partial \boldsymbol{u}} \frac{\partial h_0(\boldsymbol{x},\bar{\boldsymbol{x}},\boldsymbol{\theta})}{\partial \boldsymbol{\theta}}$. Since \eqref{prob: constraints as optimization} can be treated as an extremely simplified \eqref{prob: LQR with linear inequalities} (with $T$ reduced to 1 and $\boldsymbol{x}$ removed from $\boldsymbol{\tau}$), one can solve for the corresponding \eqref{prob: alt problem to solve for the analytical gradient} for $\partial \mathfrak{c}_{G,l}(\boldsymbol{u})/\partial \boldsymbol{u}$. For example, for upper bound constraint $\boldsymbol{u} \leq \boldsymbol{u}_{\max}$, if $\boldsymbol{u}^\star = \boldsymbol{u}_{\max}$ then $\partial \mathfrak{c}_{G,l}(\boldsymbol{u}) / \partial \boldsymbol{u} = 0$, which further indicates $\partial h / \partial \boldsymbol{\theta} = 0$.
    This observation extends our conclusion right before this remark in the sense that control saturation can degrade learning efficiency for general controllers with constraints due to zero gradient at the time of saturation.

\begin{table*}[t]
\caption{Diagonal elements for the initial parameters, the learned parameters, and the hand-tuned parameters applied in the RotorPy experiments.}
\centering
\begin{tabular}{l|r|r|r|r|r}\toprule
Weights  & $\boldsymbol{p}$                    &$\boldsymbol{v}$                    & $\boldsymbol{q}$                          & $\boldsymbol{w}$                     & $\boldsymbol{u}$                          \\ \midrule
Initial    & {[}5,5,5{]}          & {[}0.1,0.1,0.1{]}   & {[}0.1,0.1,0.1,0.1{]}     & {[}0.05,0.05,0.05{]} & {[}0.1,0.1,0.1,0.1{]}     \\
Learned    & {[}5.01,5.03,5.00{]} & {[}0.01,0.01,0.1{]} & {[}0.13,0.11,0.09,0.10{]} & {[}0.01,0.01,0.05{]} & {[}0.01,0.01,0.01,0.01{]} \\
Hand-Tuned & {[}10,10,10{]}       & {[}0.1,0.1,0.1{]}   & {[}0.1,0.1,0.1,0.1{]}     & {[}0.05,0.05,0.05{]} & {[}0.1,0.1,0.1,0.1{]} \\ 
        \bottomrule
           
\end{tabular}
\label{table:QR}
\end{table*}

\subsection{Details of the experiment setup in RotorPy}

We present details on the setup of the experiments in Section~\ref{sec:experiment}. We present the values for the initial parameters before training, the learned parameters after training, and the hand-tuned parameters in Table~\ref{table:QR}.
Regarding the hand-tuning of the controller, we started hand-tuning by setting all controller parameters to 1 and finally achieved the hand-tuned parameters displayed in Table~\ref{table:QR} after 10 trials. We originally initiated the DiffTune-MPC with parameters equal to the hand-tuned parameters. However, the tracking error attained by the auto-tuned parameters is not significantly smaller than that achieved by the hand-tuned parameters. This result implied that the hand-tuned parameters are actually close to a local minimum. Next, we set the initial parameters of DiffTune-MPC as the displayed initial parameters in Table~\ref{table:QR} by perturbing the $\boldsymbol{p}$-weights of the hand-tuned parameters. With the new initial parameters, DiffTune-MPC results in a set of learned parameters with much better tracking performance than achieved by the hand-tuned parameters.

\begin{figure}[t]
    \centering    \includegraphics[width=1\columnwidth]{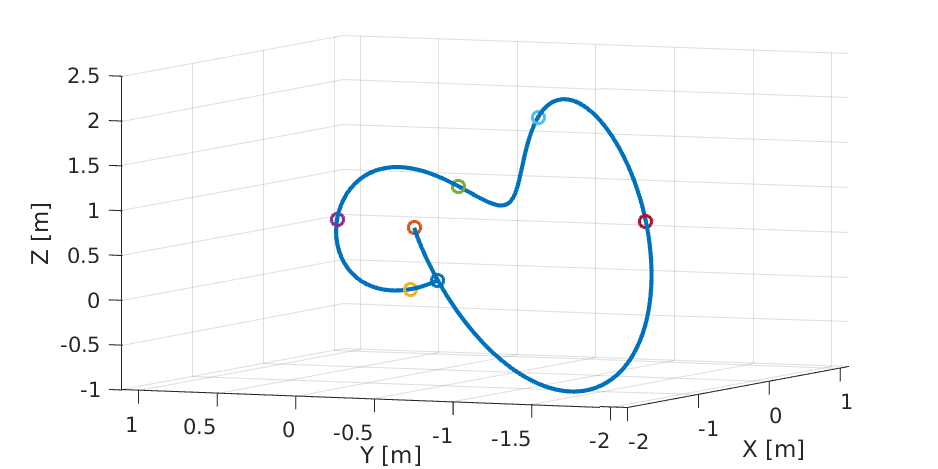} 
    \caption{Polynomial Trajectory 1 used in Table~\ref{table:comp_W}}   
\label{fig: poly_1}
\end{figure}

\begin{figure}[t]
    \centering
    \includegraphics[width=1\columnwidth]{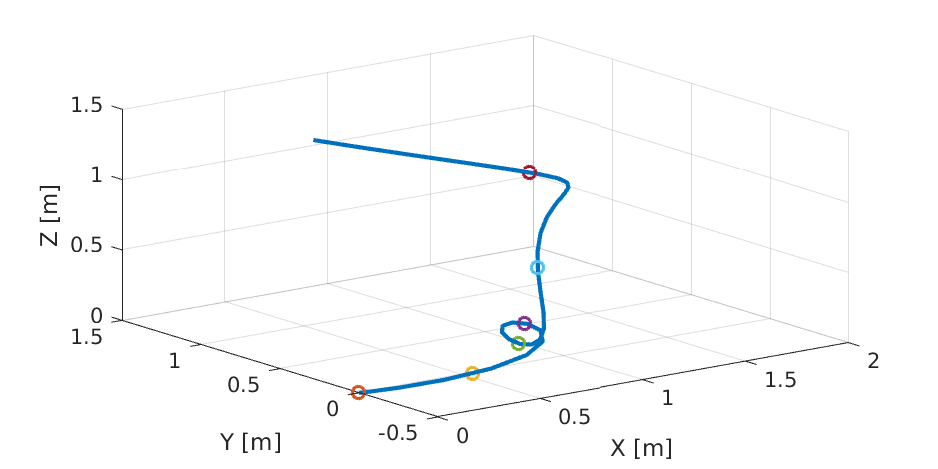} 
    \caption{Polynomial Trajectory 2 used in Table~\ref{table:comp_W}}   
\label{fig: poly_2}
\end{figure}

\begin{figure}[ht!]
    \centering
    \includegraphics[width=1\columnwidth]{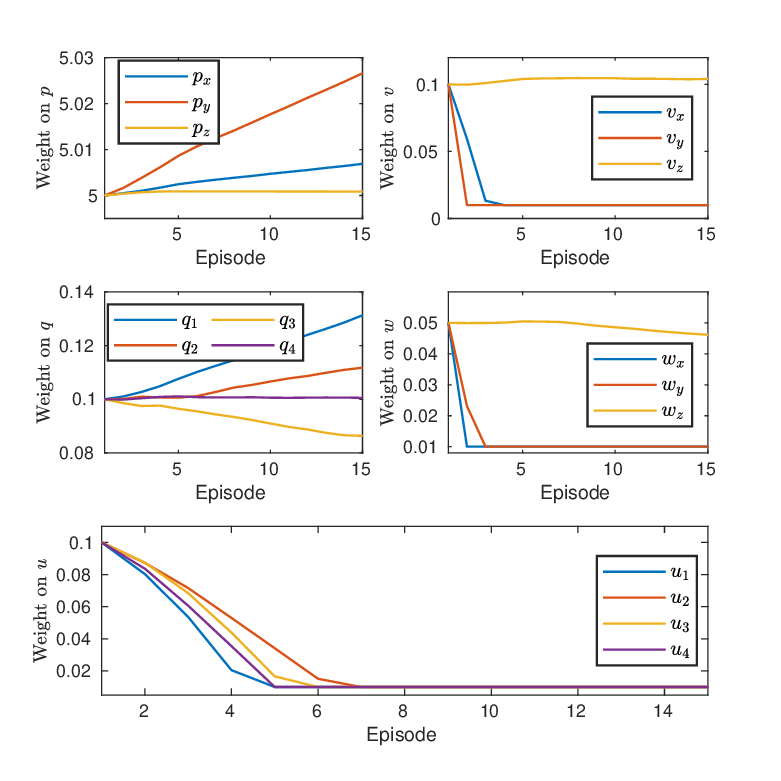} 
    \vspace{-5mm}
    \caption{Evolution of the quadrotor parameters with DiffTune-MPC used in RotorPy.}   
\label{fig: Q_R}
\vspace{-5mm}
\end{figure}

The two polynomial trajectories are plotted in Figs. \ref{fig: poly_1} and \ref{fig: poly_2}. The waypoints for generating the first polynomial trajectory are [0,0,0], [0.5,0.4,-0.2], [0.8,1,0.5], [0.3,0,1], [-0.8,-1,2], [-1.5,-2,1], [0,0,0], [1,0.6,0.4], and the average velocity of the quadrotor is set to be 1 m/s. The waypoints for generating the second polynomial trajectory are [0,0,0], [0.4,-0.2,0.1], [0.5,-0.4,0.5], [0.7,-0.1,0.2], [1.1,0.3,0.5],  [1.6,1,0.8],[0.7,1.2,1.2], and the average velocity is set to be 2 m/s.
In addition to the evolution of RMSE through the learning episodes presented in the main text, we plot the evolution of each parameter during training in Fig. \ref{fig: Q_R}. It can be observed that the weights for velocities $\boldsymbol{v}$ and angular velocities $\boldsymbol{w}$ in both $x$ and $y$ directions, as well as the weights for all control inputs all decrease to the minimum value 0.01, while the rest of the parameters do not change much.

\end{document}